\RequirePackage[svgnames,table]{xcolor}
\documentclass[10pt,letterpaper,twocolumn]{templates/kiramei/template/style}
\usepackage{stfloats,tikz}
\usepackage{algorithm,algpseudocode}

\usepackage{graphicx}
\usepackage{float,epstopdf}
\usepackage{bbm}

\usepackage{microtype}

\usepackage[numbers]{natbib}
\setcitestyle{square}

\usepackage{subcaption}
\usepackage{booktabs}

\usepackage{amsmath}
\usepackage{amssymb}
\usepackage{mathtools}
\usepackage{amsthm}
\usepackage{dsfont}
\usepackage{makecell}
\usepackage{multirow} 
\usepackage{amsfonts} 
\usepackage{mathrsfs}
\usepackage[amssymb, thickqspace]{SIunits}
\DeclareFontFamily{U}{rsfs}{\skewchar\font127}
\DeclareFontShape{U}{rsfs}{m}{n}{%
  <5> <5.5> <6> rsfs5
  <7> rsfs7
  <8> <9> <10> <10.95> <12> <14.4> <17.28> <20.74> <24.88> rsfs10
}{}
\DeclareFontShape{OML}{eur}{m}{n}{%
  <5> <5.5> <6> <7> <8> <9> gen * eurm
  <10> <10.95> <12> <14.4> <17.28> <20.74> <24.88> eurm10
}{}
\usepackage{enumitem}
\usepackage{pgfplotstable}
\pgfplotsset{compat=1.18}
\usepackage{lipsum}		

\usepackage{microtype}
\usepackage{graphicx}
\usepackage{booktabs} 
\usepackage[table]{xcolor}
\usepackage{arydshln}
\usepackage[normalem]{ulem} 

\usepackage{cases}
\usepackage{wrapfig}

\usepackage{url}

\usepackage{thmtools}
\usepackage{thm-restate}
\usepackage{tabu}

\definecolor{huskypurple}{HTML}{4B2E83}

\usepackage{titletoc}

\usepackage{listings}
\lstdefinestyle{promptstyle}{
  basicstyle=\ttfamily\footnotesize,
  breaklines=true,
  breakautoindent=false,
  breakindent=0pt,
  postbreak=\mbox{\textcolor{gray}{$\hookrightarrow$}\space},
  columns=fullflexible,
  keepspaces=true,
  frame=single,
  framesep=5pt,
  xleftmargin=6pt,
  xrightmargin=6pt,
  aboveskip=8pt,
  belowskip=8pt,
  showstringspaces=false,
}

\usepackage{fontawesome5}   

\makeatletter
\def\munderbar#1{\underline{\sbox\tw@{$#1$}\dp\tw@\z@\box\tw@}}
\makeatother

\AddToHook{cmd/appendix/before}{%
  \setcounter{axiom}{0}%
}

\newcommand{\be}{\begin{equation}}
\newcommand{\ee}{\end{equation}}
\newcommand{\bea}{\begin{equation*}\begin{aligned}}
\newcommand{\eea}{\end{aligned}\end{equation*}}

\newcommand{\monobold}[1]{{\fontfamily{lmtt}\bfseries\selectfont #1}}

\renewcommand{\headingfont}{\color{Black}\bfseries\fontsize{12}{14}\selectfont}
\titleformat{\subsection}{\bfseries\fontsize{11}{13}\selectfont}{\thesubsection.}{0.5em}{#1}[]

\title{GraSLA: A Lightweight Spatial-Language-Action Model with Grasp Hypotheses as Geometric Priors}
\author{Jibao Yuan\textsuperscript{1,*}, Zixiang Zhao\textsuperscript{1,*}, Wensong Xu\textsuperscript{1,*}, Siqi Lu\textsuperscript{1,*},\authorcr
Yinzhen Lv\textsuperscript{2}, Qingling Zhong\textsuperscript{1}, Shun Li\textsuperscript{1}, Xiaolong Su\textsuperscript{1},\authorcr
Shengqiang Li\textsuperscript{1}, Chao Xu\textsuperscript{1}, Longwen Chen\textsuperscript{1}, Shaofei Chen\textsuperscript{1,\textdagger}\\
{\normalfont\small\textsuperscript{1}College of Intelligence Science and Technology, National University of Defense Technology}\\
{\normalfont\small\textsuperscript{2}Faculty of Engineering and Information Technology, The University of Melbourne, Melbourne, Australia}\\
{\normalfont\footnotesize\textsuperscript{*}Equal contribution; \textsuperscript{\textdagger}Corresponding author}\\
{\normalfont\footnotesize\faGithub \textbf{Source Code:} \href{https://github.com/Minakanmi-Yuki/GraSLA}{\monobold{https://github.com/Minakanmi-Yuki/GraSLA}}}}
\runningtitle{GraSLA: Grasp Priors for Manipulation}
\hypersetup{pdftitle={GraSLA: A Lightweight Spatial-Language-Action Model with Grasp Hypotheses as Geometric Priors}}

\usepackage{pifont}
\definecolor{GraSLARed}{HTML}{C12F32}
\definecolor{GraSLAGreen}{HTML}{217F45}
\definecolor{GraSLABlue}{HTML}{2863AA}
\newcommand{\GraSLARankOne}[1]{\textcolor{GraSLARed}{\textbf{#1}}}
\newcommand{\GraSLARankTwo}[1]{\textcolor{GraSLAGreen}{\textbf{#1}}}
\newcommand{\GraSLARankThree}[1]{\textcolor{GraSLABlue}{\textbf{#1}}}
\newcommand{\GraSLACheck}{\textcolor{GraSLAGreen}{\ding{51}}}
\newcommand{\GraSLACross}{\textcolor{GraSLARed}{\ding{55}}}
\makeatletter
\newcommand{\GraSLATableHighlight}{\rlap{\color{black!10}\rule[-\dp\@arstrutbox]{\textwidth}{\dimexpr\ht\@arstrutbox+\dp\@arstrutbox\relax}}}
\newcommand{\GraSLAColumnHighlight}{\rlap{\color{black!10}\rule[-\dp\@arstrutbox]{\columnwidth}{\dimexpr\ht\@arstrutbox+\dp\@arstrutbox\relax}}}
\global\let\GraSLADividerNext\relax
\newcommand{\GraSLAMarkTop}[1]{\gdef\GraSLADividerNext{\tikz[remember picture,overlay,baseline]{\coordinate (gra-#1-top) at (0,\ht\@arstrutbox);}}}
\newcommand{\GraSLAMarkBottom}[1]{\gdef\GraSLADividerNext{\tikz[remember picture,overlay,baseline]{\draw[black,line width=0.4pt] (gra-#1-top) -- (0,-\dp\@arstrutbox);}}}
\newcommand{\GraSLATableDivider}{\GraSLADividerNext\global\let\GraSLADividerNext\relax\hspace{0.4pt}}
\makeatother
\begin{document}
\flushbottom
\addtolength{\parskip}{0pt plus 1pt minus 0.5pt}
\addtolength{\bibsep}{0pt plus 1pt minus 1pt}
\begin{abstract}
This paper focuses on improving the precision and adaptability of language-conditioned manipulation policies learned through imitation learning. Recently, action-based methods have become mainstream for closed-loop control, yet implicit behavior learning from demonstrations limits their generalization beyond training conditions. To address this, explicit geometric priors offer direct guidance for action prediction. In particular, 6-DoF grasp detection generates diverse grasp hypotheses with explicit approach directions for the gripper. Yet raw hypotheses span multiple regions and lack task semantics, requiring joint reasoning with the scene and instruction. To this end, we propose GraSLA, a lightweight spatial-language-action model with grasp hypotheses as explicit geometric priors. Specifically, it encodes multi-view point clouds, the instruction, and grasp hypotheses, binds task-relevant ones to scene context and task semantics via bidirectional cross-attention, and decodes continuous action chunks. Experiments show that, with 0.29B parameters, 39.5 ms latency, and no robot-data pretraining, GraSLA outperforms existing baselines, achieving 99.3\% on LIBERO, 73.7\% on LIBERO+, and 53.33\% on LIBERO-Pro.
\end{abstract}
\makeatletter
\twocolumn[
{\titlefont\@title\par}
\vspace{0.4em}
{\centering\@author\par}
\vspace{0.7em}
\vspace{0.5em}
]
\makeatother

\begin{center}
{\large\bfseries Abstract}
\end{center}
{\absfont\itshape\noindent\theabstract\par}

\section{Introduction}
Robotic manipulation encompasses diverse physical interaction tasks across industrial, warehouse, and household settings \citep{industrial-insertion,warehouse-robin,rt1}. Powered by vision-language models, modern manipulation policies can follow natural-language instructions and have achieved promising performance across a range of tasks \citep{openvla,pi0,openvlaoft}.

\begin{figure}[t]
\centering
\includegraphics[width=\columnwidth]{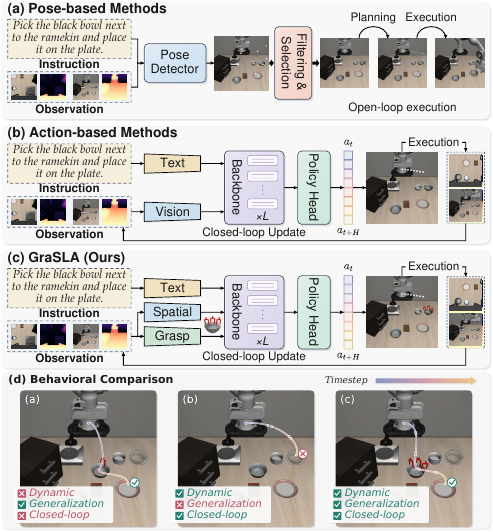}
\caption{Comparison of manipulation paradigms and behavior under position perturbations. (a)--(c) Execution pipelines of existing paradigms and GraSLA, which incorporates grasp hypotheses into closed-loop control. (d) Behavioral comparison when the target shifts from its original position.}
\label{fig:teaser}
\end{figure}

Despite this progress, translating instructions into precise actions remains challenging. To do so, robots must relate the instruction to the scene and use geometry to guide gripper interactions \citep{spatialvla,diffuseractor}. Focusing on parallel-jaw manipulation, existing methods fall into two paradigms, as shown in Fig.~\ref{fig:teaser}(a) and Fig.~\ref{fig:teaser}(b) \citep{ppi}: (1) \textit{pose-based methods}, which predict target end-effector poses for open-loop motion planning \citep{rvt2,peract}; and (2) \textit{action-based methods}, which directly predict continuous actions from current observations for closed-loop control \citep{pi0,diffusionpolicy}. The former provide explicit geometric grounding for spatial generalization, whereas the latter enable closed-loop feedback for reactive control.

However, both paradigms face distinct limitations. Pose-based methods struggle to react to dynamic environments during open-loop execution \citep{graspldp}. Conversely, action-based methods learn manipulation behaviors implicitly from demonstrations, which often limits their generalization beyond training conditions \citep{libero-plus,libero-pro}. As shown in Fig.~\ref{fig:teaser}(d), an action-based policy still moves toward the original position when the target shifts. A natural remedy is therefore to incorporate explicit geometric priors into closed-loop action prediction \citep{spatialrobograsp}.

In particular, 6-DoF grasp detection generates grasp hypotheses that can serve as such priors, offering explicit approach directions for the gripper \citep{contactgraspnet,graspness}. Yet raw hypotheses span multiple scene regions and lack task semantics, so their relevance depends on both the scene and the instruction \citep{graspgpt,m2t2}. Rather than relying on heuristic grasp selection, keeping a diverse set of hypotheses preserves valid options. This motivates us to integrate these diverse hypotheses with spatial and language representations, identifying task-relevant geometric priors for action prediction.

To this end, as illustrated in Fig.~\ref{fig:teaser}(c), we propose \textit{GraSLA}, a lightweight spatial-language-action (SLA) model that uses grasp hypotheses as geometric priors. Specifically, GraSLA first encodes multi-view point clouds with Utonia \citep{utonia}, the instruction with BERT \citep{bert}, and grasp hypotheses from GSNet \citep{graspness} into spatial, language, and grasp tokens, respectively. Next, stacked SLG interaction blocks fuse these streams via bidirectional cross-attention, binding task-relevant hypotheses to the scene context and task semantics. Finally, an ACT-style decoder \citep{act} predicts an action chunk from the fused memory and robot state.

Our contributions can be summarized as follows:

\begin{itemize}
\item We propose GraSLA, a lightweight spatial-language-action model that incorporates \textit{grasp hypotheses as explicit geometric priors}, achieving efficient inference and strong generalization.
\item We incorporate a pretrained 6-DoF grasp detector and design \textit{spatial-language-grasp} (SLG) interaction blocks, retaining diverse grasp hypotheses and binding task-relevant hypotheses to the scene and the instruction via bidirectional cross-attention for action prediction.
\item In simulation, GraSLA achieves success rates of 99.3\% on LIBERO, 73.7\% on LIBERO+, and 53.3\% on LIBERO-Pro, demonstrating strong performance under standard, perturbed, and shifted conditions.
\end{itemize}

\section{Related Work}
\textbf{3D for Manipulation.} Beyond 2D visuomotor policies \citep{act,diffusionpolicy}, 3D methods leverage geometry to resolve depth ambiguity and improve generalization \citep{diffuseractor}. Early 3D methods represent scenes with voxels, rendered views, or 3D feature fields to ground manipulation in 3D space \citep{peract,rvt2,act3d}. To avoid dense reconstruction, subsequent methods instead encode point clouds directly \citep{polarnet,dp3,rise,ppi,pointvla,geovla,pointact,any3dvla}, while point-cloud pretraining provides transferable representations for downstream manipulation \citep{sonata,utonia}. By contrast, recent vision-language-action (VLA) models retain 2D backbones and augment them via spatial encodings \citep{spatialvla,threedsvla} or 3D feature alignment \citep{spatialforcing,sla} to bridge semantics and geometry. However, these methods focus on scene structure rather than explicit geometric priors for gripper interaction, leaving feasible contact poses to be learned implicitly from demonstrations.

\begingroup
\clubpenalty=10000\relax
\textbf{Grasp-Guided Manipulation.} 6-DoF grasp detection predicts grasp poses from point clouds to provide explicit approach directions for the gripper \citep{graspnet,contactgraspnet,graspness,anygrasp,economicgrasp,grare}. To use grasp priors in policy learning, some methods fuse implicit features from grasp-pretrained backbones \citep{gparam}, but they lose explicit 6-DoF geometry. Others predict intermediate grasp poses before action prediction \citep{graspvla}, yet require grasp supervision during training. Closer to our setup, external-detection methods condition policies on a single 2D-lifted grasp \citep{spatialrobograsp} or a heuristically selected 6-DoF pose \citep{graspldp}. However, committing to a single pose discards diverse grasp hypotheses and ignores task semantics in cluttered, language-conditioned scenes.
\par
\endgroup

\textbf{Connections and Differences.} GraSLA bridges both lines by combining pretrained 3D representations \citep{utonia} with a frozen 6-DoF grasp detector \citep{graspness}. However, unlike methods that rely on implicit features or grasp supervision \citep{gparam,graspvla}, we directly inject explicit 6-DoF hypotheses without grasp annotations. Rather than relying on single-pose heuristics \citep{spatialrobograsp,graspldp}, GraSLA also preserves diverse hypotheses and binds them to scene context and task semantics via SLG interaction blocks. Thus, grasp priors extend beyond grasping to support general closed-loop manipulation.

\section{Preliminaries}
\textbf{6-DoF Grasp Detection.} Given a scene point cloud, a 6-DoF grasp detector predicts a set of grasp hypotheses $\{g_i\}_{i=1}^{N_g}$, where $N_g$ is the number of hypotheses. Each hypothesis is represented as
\begin{equation}
g_i=(T_i,w_i,h_i,d_i,q_i), \qquad T_i=(R_i,\mathbf p_i)\in SE(3),
\end{equation}
where $T_i$ specifies the gripper pose, with rotation $R_i$ and position $\mathbf p_i$. The attributes $w_i$, $h_i$, $d_i$, and $q_i$ denote the opening width, gripper height, grasp depth, and predicted grasp quality, respectively.

\textbf{Policy Formulation.} At time $t$, let $\mathcal P_t=\{\mathcal P_t^{(v)}\}_{v=1}^{N_v}$, $\ell_t$, and $\mathbf{x}_t$ denote the multi-view point clouds, language instruction, and robot state, respectively, and let $\mathcal G_t$ denote the sampled grasp hypotheses along with their local features and collision states. The policy $\pi_\theta$ predicts an $H$-step action chunk:
\begin{equation}
\widehat{\mathbf A}_t
=\pi_\theta(\mathcal P_t,\ell_t,\mathbf{x}_t,\mathcal G_t)
\in\mathbb R^{H\times d_a},
\end{equation}
where $d_a$ is the action dimension. We train the policy by behavior cloning on expert demonstrations.

\section{Method}
\begin{figure*}[!t]
\centering
\includegraphics[width=\textwidth]{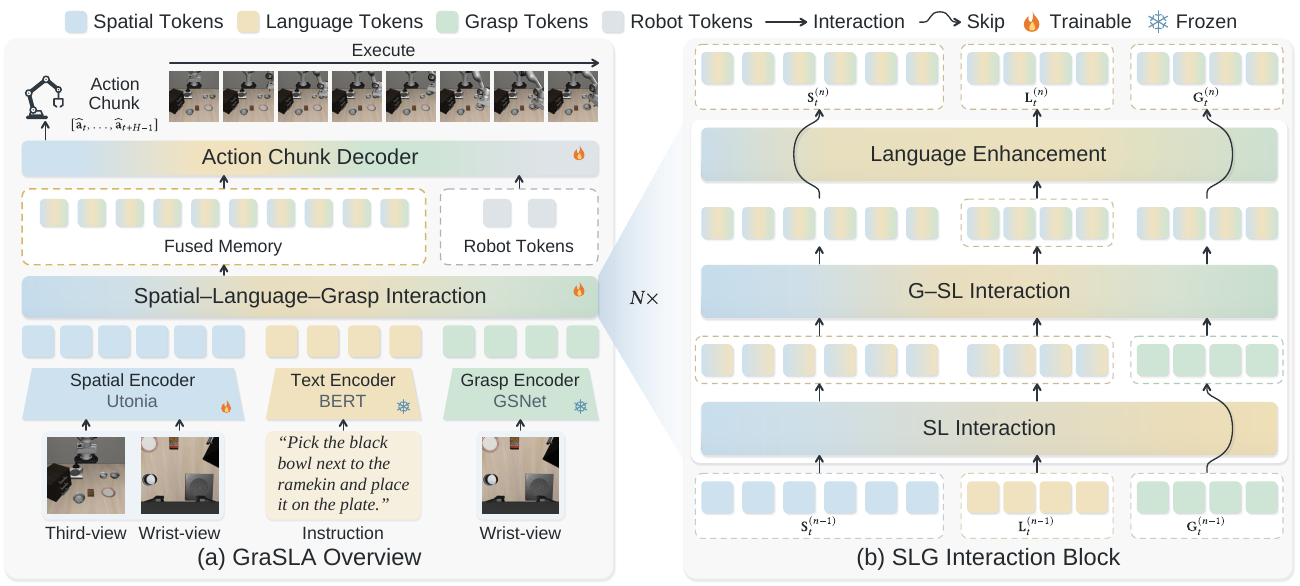}
\caption{Overview of GraSLA. (a) Multi-view point clouds, the language instruction, and grasp hypotheses are encoded into three token streams, fused via stacked SLG blocks, and decoded with the robot state into an action chunk. (b) Inside each SLG block, bidirectional cross-attention first mutually aligns the spatial and language tokens, and then jointly updates them with the grasp tokens to exchange task and geometric cues, followed by language refinement.}
\label{fig:overview}
\end{figure*}

Fig.~\ref{fig:overview} illustrates the overall design of GraSLA. Directly feeding raw 6-DoF grasp hypotheses into a policy is ineffective, as they are redundant, task-agnostic, and must adapt to the dynamic scene and instruction. To bridge this gap, GraSLA operates in three stages: (1) preserving diverse, collision-checked hypotheses during encoding (Sec.~\ref{sec:encoding}), (2) binding task-relevant hypotheses to the scene and instruction via SLG interaction (Sec.~\ref{sec:slg}), and (3) decoding the fused memory with the robot state into action chunks (Sec.~\ref{sec:action}). Model details are provided in Appendix \ref{app:training}.

\subsection{Multimodal Encoding}
\label{sec:encoding}
Effective grasp-guided control requires representing the scene, the instruction, and the geometric priors in a common token space. In the spatial branch, a shared Utonia backbone \citep{utonia} encodes the point cloud $\mathcal P_t^{(v)}$ of each view $v$. Its multi-level features are concatenated and downsampled via farthest point sampling (FPS) into $\mathbf F_t^{(v)}$, yielding the spatial tokens
\begin{equation}
\mathbf S_t^{(0)}
=\operatorname{Concat}_{v=1}^{N_v}
\left[\phi_S\!\left(\mathbf F_t^{(v)}\right)+\mathbf e_v\right],
\label{eq:scene-token}
\end{equation}
where $\phi_S$ is a learnable projection and $\mathbf e_v$ is a learnable view embedding. In the language branch, BERT \citep{bert} encodes the instruction $\ell_t$ into final-layer hidden states $\mathbf F_t^L$, giving the language tokens
\begin{equation}
\mathbf L_t^{(0)}=\phi_L(\mathbf F_t^L),
\label{eq:language-token}
\end{equation}
where $\phi_L$ is a learnable linear projection. In the grasp branch, GSNet \citep{graspness} predicts raw hypotheses $\{g_i\}_{i=1}^{N_g}$ from the wrist-view point cloud. To prune redundant hypotheses while preserving object coverage and feasibility, we evaluate their collision states $\mathbf z_t$ and sample across regions into $\mathcal G_t$, giving the grasp tokens
\begin{equation}
\mathbf G_t^{(0)}=\phi_G(\mathcal G_t)+\mathbf e_z(\mathbf z_t),
\label{eq:grasp-token}
\end{equation}
where $\phi_G$ is a learnable GraspEncoder and $\mathbf e_z$ is a learnable collision-state embedding. All sampled tokens are retained so that subsequent fusion (Sec.~\ref{sec:slg}) can identify task-relevant hypotheses.

Finally, the robot state $\mathbf{x}_t$ is encoded for action prediction, yielding the state tokens
\begin{equation}
\mathbf X_t=\operatorname{LN}\!\left(\phi_R(\mathbf{x}_t)+\mathbf e_R\right),
\label{eq:robot-token}
\end{equation}
where $\phi_R$ is a learnable state projection, $\mathbf e_R$ is a learnable robot embedding, and $\operatorname{LN}$ denotes layer normalization.

\subsection{Multimodal Fusion}
\label{sec:slg}
To fuse the three token streams, GraSLA stacks $N$ spatial-language-grasp (SLG) interaction blocks (Fig.~\ref{fig:overview}(b)), updating the tokens as
\begin{equation}
\begin{aligned}
&(\mathbf S_t^{(n)},\mathbf L_t^{(n)},\mathbf G_t^{(n)})\\
&\quad=\operatorname{SLG}_n
\bigl(\mathbf S_t^{(n-1)},\mathbf L_t^{(n-1)},\mathbf G_t^{(n-1)}\bigr),
\end{aligned}
\label{eq:fusion}
\end{equation}
where $\operatorname{SLG}_n$ denotes the $n$-th interaction block. Each block operates in three stages. First, to infuse task semantics into the spatial tokens and scene context into the language tokens, the two streams interact via bidirectional cross-attention:

\begin{algorithm}[t]
\small
\captionsetup{labelsep=space}
\caption{Training and inference of GraSLA}
\label{alg:training-overview}
\algrenewcommand{\algorithmicrequire}{\textbf{Input}}
\algrenewcommand{\algorithmicensure}{\textbf{Output}}
\algrenewcommand\alglinenumber[1]{\footnotesize #1}
\algrenewcommand{\algorithmicindent}{1em}
\begin{algorithmic}[1]
\Require Demonstrations $\mathcal D$, horizon $H$, $N$ SLG blocks
\Ensure Policy $\pi_\theta$
\Statex \textbf{Training}
\State Initialize $\theta$ and freeze BERT and GSNet.
\While{training is not complete}
    \State Sample a batch $(\mathcal P_t$, $\ell_t$, $\mathbf x_t$, $\mathbf A_t^\star$, $\mathbf m_t)$ from $\mathcal D$.
    \State Obtain $\mathcal G_t$ using GSNet on wrist-view point clouds.
    \State Encode inputs into $\mathbf S_t^{(0)}$, $\mathbf L_t^{(0)}$, $\mathbf G_t^{(0)}$, and $\mathbf X_t$.
    \State Fuse $\mathbf S_t^{(0)}$, $\mathbf L_t^{(0)}$, and $\mathbf G_t^{(0)}$ into $\mathbf M_t$ via $N$ SLG blocks.
    \State Predict $\widehat{\mathbf A}_t$ from $\mathbf M_t$ and $\mathbf X_t$.
    \State Update $\theta$ using the masked $\ell_1$ loss $\mathcal L_{\mathrm{BC}}$.
\EndWhile
\Statex \textbf{Inference}
\While{the task is not complete}
    \State Observe $\mathcal P_t$, $\ell_t$, and $\mathbf x_t$, and regenerate $\mathcal G_t$ with GSNet.
    \State Predict $\widehat{\mathbf A}_t$ with $\pi_\theta$ and execute the denormalized chunk.
\EndWhile
\end{algorithmic}
\end{algorithm}

\begin{table*}[!t]
\centering\fontsize{9}{10.8}\selectfont
\captionsetup{name=Table}
\newcommand{\GraSLATableHeading}[1]{{\fontsize{9}{10.8}\selectfont\bfseries #1}}
\newcommand{\GraSLATableLabel}[1]{{\fontsize{8.5}{10.2}\selectfont #1}}
\newcommand{\GraSLATableGroup}[1]{{\fontsize{9}{10.8}\selectfont\itshape #1}}
\setlength{\tabcolsep}{2pt}
\renewcommand{\arraystretch}{1.05}
\caption{Comparison on LIBERO. PT. denotes pretraining on additional robot demonstrations. Params denotes total parameters. Latency denotes end-to-end inference time from multimodal inputs to an action chunk or an equivalent sequence of autoregressive action tokens. VRAM usage and latency are measured on a single NVIDIA RTX 5090 GPU with batch one. The latency protocol is described in Appendix~\ref{app:latency}. The top three results in each column are shown in \GraSLARankOne{red}, \GraSLARankTwo{green}, and \GraSLARankThree{blue}.}
\label{tab:main}
\begin{minipage}{\textwidth}
\centering
\begin{tabular*}{\textwidth}{@{\extracolsep{\fill}}l!{\GraSLATableDivider}cccccccccc@{}}
\toprule
\GraSLAMarkTop{libero}\multirow{2}{*}{\GraSLATableHeading{Method}} & \multirow{2}{*}{\GraSLATableHeading{Reference}} & \multirow{2}{*}{\GraSLATableHeading{PT.}} & \multicolumn{3}{c}{\GraSLATableHeading{Efficiency (B/GiB/ms) $\downarrow$}} & \multicolumn{5}{c}{\GraSLATableHeading{LIBERO Success Rate (\%) $\uparrow$}}\\
\cmidrule(lr){4-6}\cmidrule(l){7-11}
 &  &  & \GraSLATableLabel{Params} & \GraSLATableLabel{VRAM} & \GraSLATableLabel{Latency} & \GraSLATableLabel{\textit{Spatial}} & \GraSLATableLabel{\textit{Object}} & \GraSLATableLabel{\textit{Goal}} & \GraSLATableLabel{\textit{Long}} & \GraSLATableLabel{Avg.}\\
\midrule
\multicolumn{11}{@{}l}{\GraSLATableGroup{Large-scale}}\\
OpenVLA \cite{openvla} & CoRL~24 & \GraSLACheck & 7.5 & 15.1 & 140.1 & 84.7 & 88.4 & 79.2 & 53.7 & 76.5\\
OpenVLA-OFT \cite{openvlaoft} & RSS~25 & \GraSLACheck & 7.7 & 15.8 & 101.2 & 97.6 & 98.4 & 97.9 & 94.5 & 97.1\\
$\pi_{0.5}$ \cite{pi05} & CoRL~25 & \GraSLACheck & 3.4 & 24.1 & 61.1 & 98.8 & 98.2 & 98.0 & 92.4 & 96.9\\
UniVLA \cite{univla} & RSS~25 & \GraSLACheck & 7.6 & 15.3 & 98.4 & 96.5 & 96.8 & 95.6 & 92.0 & 95.2\\
CogVLA \cite{cogvla} & NeurIPS~25 & \GraSLACheck & 8.5 & 18.2 & 65.9 & 98.6 & 98.8 & 96.6 & 95.4 & 97.4\\
Spatial Forcing \cite{spatialforcing} & ICLR~26 & \GraSLACheck & 7.7 & 16.1 & 95.5 & \GraSLARankTwo{99.4} & \GraSLARankThree{99.6} & 98.8 & 96.0 & 98.5\\
VLA-0 \cite{vla0} & arXiv~25 & \GraSLACross & 3.8 & 8.2 & 1777.1 & 97.0 & 97.8 & 96.2 & 87.6 & 94.7\\
PointmapVLA \cite{pointmapvla} & Report~26 & \GraSLACheck & 6.1 & 11.6 & 95.7 & \GraSLARankTwo{99.4} & 99.5 & \GraSLARankTwo{99.3} & \GraSLARankThree{97.9} & \GraSLARankThree{99.0}\\
Flex-$\pi^\ast$ \cite{flexpi} & arXiv~26 & \GraSLACheck & 6.0 & 24.7 & 590.7 & \GraSLARankThree{99.2} & \GraSLARankTwo{99.8} & \GraSLARankThree{99.0} & \GraSLARankTwo{98.6} & \GraSLARankTwo{99.2}\\
\midrule
\multicolumn{11}{@{}l}{\GraSLATableGroup{Lightweight}}\\
TurboVLA \cite{turbovla} & arXiv~26 & \GraSLACross & \GraSLARankOne{0.2} & \GraSLARankOne{1.0} & \GraSLARankOne{12.0} & 97.0 & \GraSLARankOne{100.0} & 98.2 & 95.2 & 97.6\\
SmolVLA \cite{smolvla} & arXiv~25 & \GraSLACross & \GraSLARankThree{0.5} & \GraSLARankTwo{1.5} & 83.5 & 90.0 & 96.0 & 92.0 & 71.0 & 87.3\\
DreamVLA \cite{dreamvla} & NeurIPS~25 & \GraSLACheck & 0.7 & 5.7 & \GraSLARankTwo{37.2} & 97.5 & 94.0 & 89.5 & 89.5 & 92.6\\
Evo-Depth \cite{evodepth} & NeurIPS~26 & \GraSLACross & 0.9 & 3.4 & 170.2 & 95.6 & 99.2 & 95.6 & 91.3 & 95.4\\
X-VLA \cite{xvla} & ICLR~26 & \GraSLACheck & 0.9 & 4.2 & 78.4 & 98.2 & 98.6 & 97.8 & 97.6 & 98.1\\
VLA-Adapter-Pro \cite{vlaadapter} & AAAI~26 & \GraSLACross & 1.5 & 5.4 & 43.2 & \GraSLARankOne{99.6} & \GraSLARankThree{99.6} & 98.2 & 96.4 & 98.5\\
\GraSLAMarkBottom{libero}\GraSLATableHighlight \textit{GraSLA (ours)} & --- & \GraSLACross & \GraSLARankTwo{0.29} & \GraSLARankThree{3.1} & \GraSLARankThree{39.5} & \GraSLARankOne{99.6} & 99.0 & \GraSLARankOne{99.8} & \GraSLARankOne{98.8} & \GraSLARankOne{99.3}\\
\bottomrule
\end{tabular*}
\end{minipage}
\end{table*}
\begin{equation}
\bigl(\widetilde{\mathbf S}_t^{(n)},\widetilde{\mathbf L}_t^{(n)}\bigr)
=\operatorname{BiCA}\bigl(\mathbf S_t^{(n-1)},\mathbf L_t^{(n-1)}\bigr),
\label{eq:sl-interaction}
\end{equation}
where $\operatorname{BiCA}$ denotes bidirectional cross-attention. Next, to focus the grasp tokens on task-relevant hypotheses while injecting approach geometry into the spatial and language tokens, the grasp tokens and the concatenated spatial-language tokens jointly update each other via $\operatorname{BiCA}$:
\begin{equation}
\begin{aligned}
&\bigl(\mathbf G_t^{(n)},[\mathbf S_t^{(n)};\widehat{\mathbf L}_t^{(n)}]\bigr)\\
&\quad=\operatorname{BiCA}\bigl(\mathbf G_t^{(n-1)},
[\widetilde{\mathbf S}_t^{(n)};\widetilde{\mathbf L}_t^{(n)}]\bigr),
\end{aligned}
\label{eq:g-sl-interaction}
\end{equation}
where $[\cdot;\cdot]$ denotes token concatenation. Finally, to consolidate the instruction representation after cross-modal fusion, the language tokens are refined as
\begin{equation}
\mathbf L_t^{(n)}=\operatorname{LE}\bigl(\widehat{\mathbf L}_t^{(n)}\bigr),
\label{eq:language-enhancement}
\end{equation}
where $\operatorname{LE}$ denotes the language enhancement module, comprising self-attention and a feed-forward network. After $N$ blocks, the three streams are concatenated into the fused memory for action prediction (Sec.~\ref{sec:action}):
\begin{equation}
\mathbf M_t=[\mathbf S_t^{(N)};\mathbf L_t^{(N)};\mathbf G_t^{(N)}].
\label{eq:fusion-memory}
\end{equation}

\subsection{Action Prediction}
\label{sec:action}
Based on the fused memory $\mathbf M_t$ and state tokens $\mathbf X_t$, an ACT-style decoder \citep{act} predicts an $H$-step action chunk:
\begin{equation}
\widehat{\mathbf A}_t
=\tanh\!\left(\phi_A\!\left(
\operatorname{Dec}\bigl(\mathbf Q_H,[\mathbf M_t;\mathbf X_t]\bigr)
\right)\right),
\label{eq:action-head}
\end{equation}
where $\mathbf Q_H$ denotes $H$ learnable action queries and $\phi_A$ is a learnable action projection. During training, the policy is supervised with a masked $\ell_1$ loss between $\widehat{\mathbf A}_t$ and the expert action chunk $\mathbf A_t^\star$. Algorithm~\ref{alg:training-overview} summarizes the training and inference procedure.

\section{Experiments}
In our experiments, we evaluate the task performance and inference efficiency of GraSLA and examine the role of grasp hypotheses. The study covers single-arm manipulation on LIBERO \citep{libero} and ablation studies.

\begin{table*}[t]
\centering\fontsize{9}{10.8}\selectfont
\captionsetup{name=Table}
\newcommand{\GraSLATableHeading}[1]{{\fontsize{9}{10.8}\selectfont\bfseries #1}}
\newcommand{\GraSLATableLabel}[1]{{\fontsize{8.5}{10.2}\selectfont #1}}
\newcommand{\GraSLATableGroup}[1]{{\fontsize{9}{10.8}\selectfont\itshape #1}}
\setlength{\tabcolsep}{2pt}
\renewcommand{\arraystretch}{1.05}
\setlength{\tabcolsep}{1pt}
\caption{Comparison on LIBERO+. PT. denotes pretraining on additional robot demonstrations. MVUCF uses GR00T-N1.6. The top three results in each column are shown in \GraSLARankOne{red}, \GraSLARankTwo{green}, and \GraSLARankThree{blue}.}
\label{tab:liberoplus-main}
\begin{tabular*}{\textwidth}{@{\extracolsep{\fill}}l!{\GraSLATableDivider}ccccccccccc@{}}
\toprule
\GraSLAMarkTop{liberoplus}\multirow{2}{*}{\GraSLATableHeading{Method}} & \multirow{2}{*}{\GraSLATableHeading{Reference}} & \multirow{2}{*}{\GraSLATableHeading{PT.}} & \multirow{2}{*}{\GraSLATableHeading{Params (B) $\downarrow$}} & \multicolumn{8}{c}{\GraSLATableHeading{LIBERO+ Success Rate (\%) $\uparrow$}}\\
\cmidrule(l){5-12}
 &  &  &  & \GraSLATableLabel{\textit{Cam.}} & \GraSLATableLabel{\textit{Robot}} & \GraSLATableLabel{\textit{Lang.}} & \GraSLATableLabel{\textit{Light}} & \GraSLATableLabel{\textit{Bkg.}} & \GraSLATableLabel{\textit{Noise}} & \GraSLATableLabel{\textit{Layout}} & \GraSLATableLabel{Overall}\\
\midrule
\multicolumn{12}{@{}l}{\GraSLATableGroup{Large-scale}}\\
OpenVLA \cite{openvla,libero-plus} & CoRL~24 & \GraSLACheck & 7.54 & 0.8 & 3.5 & 23.0 & 8.1 & 34.8 & 15.2 & 28.5 & 15.6\\
OpenVLA-OFT$_m$ \cite{openvlaoft,libero-plus} & RSS~25 & \GraSLACheck & 7.7 & 55.6 & 21.7 & \GraSLARankTwo{81.0} & \GraSLARankThree{92.7} & 91.0 & 78.6 & 68.7 & 67.9\\
UniVLA \cite{univla,libero-plus} & RSS~25 & \GraSLACheck & 7.6 & 1.8 & 46.2 & 69.6 & 69.0 & 81.0 & 21.2 & 31.9 & 42.9\\
$\pi_{0}$ \cite{pi0,libero-plus} & RSS~25 & \GraSLACheck & 3.2 & 13.8 & 6.0 & 58.8 & 85.0 & 81.4 & \GraSLARankThree{79.0} & 68.9 & 53.6\\
$\pi_{0}$-FAST \cite{pi0fast,libero-plus} & RSS~25 & \GraSLACheck & 3.0 & \GraSLARankTwo{65.1} & 21.6 & 61.0 & 73.2 & 73.2 & 74.4 & 68.8 & 61.6\\
MemoryVLA (joint) \cite{memoryvla,memoryvlapp} & ICLR~26 & \GraSLACheck & 8.38 & 42.7 & 44.9 & \GraSLARankOne{84.4} & \GraSLARankTwo{92.8} & \GraSLARankTwo{95.0} & 62.1 & \GraSLARankOne{84.7} & \GraSLARankTwo{70.2}\\
MVUCF (N1.6) \cite{mvucf} & arXiv~26 & \GraSLACheck & 3.29 & 24.1 & \GraSLARankTwo{54.1} & 72.9 & \GraSLARankOne{95.1} & \GraSLARankThree{91.8} & 61.3 & \GraSLARankTwo{74.9} & 65.2\\
GR00T-N1.6 \cite{grootn16,mvucf} & Report~25 & \GraSLACheck & 3.29 & 20.9 & 40.2 & 35.0 & 65.4 & 76.3 & 27.8 & 51.3 & 42.8\\
NORA \cite{nora,libero-plus} & arXiv~25 & \GraSLACheck & 3.76 & 2.2 & 37.0 & 65.1 & 45.7 & 58.6 & 12.8 & 62.1 & 39.0\\
WorldVLA \cite{worldvla,libero-plus} & arXiv~25 & \GraSLACross & 7.08 & 0.1 & 27.9 & 41.6 & 43.7 & 17.1 & 10.9 & 38.0 & 25.0\\
\midrule
\multicolumn{12}{@{}l}{\GraSLATableGroup{Lightweight}}\\
DreamVLA \cite{dreamvla,unimpa} & NeurIPS~25 & \GraSLACheck & \GraSLARankThree{0.7} & \GraSLARankThree{65.0} & 40.8 & 63.5 & 85.7 & 82.6 & \GraSLARankTwo{84.9} & 74.0 & \GraSLARankThree{69.9}\\
Evo-1 \cite{evo1,evo1plus} & CVPR~26 & \GraSLACross & 0.77 & 44.78 & 49.42 & 62.91 & 83.27 & 85.32 & 70.21 & 62.82 & 63.81\\
X-VLA \cite{xvla,holobrain} & ICLR~26 & \GraSLACheck & 0.88 & 22.2 & \GraSLARankOne{87.8} & 73.1 & 88.2 & \GraSLARankOne{95.3} & 61.8 & 70.7 & 69.7\\
TLC-DiT \cite{tlcdit} & arXiv~26 & \GraSLACross & \GraSLARankTwo{0.29} & 48.78 & 37.35 & \GraSLARankThree{73.52} & 68.65 & 58.55 & 50.97 & 67.02 & 57.24\\
Multitask DiT \cite{tlcdit} & arXiv~26 & \GraSLACross & \GraSLARankOne{0.25} & 42.71 & 34.65 & 72.54 & 68.13 & 52.88 & 46.41 & 65.44 & 54.07\\
\GraSLAMarkBottom{liberoplus}\GraSLATableHighlight \textit{GraSLA (ours)} & --- & \GraSLACross & \GraSLARankTwo{0.29} & \GraSLARankOne{81.6} & \GraSLARankThree{49.9} & 69.3 & 79.1 & 77.4 & \GraSLARankOne{86.4} & \GraSLARankThree{74.2} & \GraSLARankOne{73.7}\\
\bottomrule
\end{tabular*}
\end{table*}

\begin{table*}[t]
\centering\fontsize{9}{10.8}\selectfont
\captionsetup{name=Table}
\newcommand{\GraSLATableHeading}[1]{{\fontsize{9}{10.8}\selectfont\bfseries #1}}
\newcommand{\GraSLATableLabel}[1]{{\fontsize{8.5}{10.2}\selectfont #1}}
\newcommand{\GraSLATableGroup}[1]{{\fontsize{9}{10.8}\selectfont\itshape #1}}
\setlength{\tabcolsep}{2pt}
\renewcommand{\arraystretch}{1.05}
\setlength{\tabcolsep}{1pt}
\caption{Comparison on LIBERO-Pro. PT. denotes pretraining on additional robot demonstrations. The top three results in each column are shown in \GraSLARankOne{red}, \GraSLARankTwo{green}, and \GraSLARankThree{blue}.}
\label{tab:liberopro-main}
\begin{tabular*}{\textwidth}{@{\extracolsep{\fill}}l!{\GraSLATableDivider}cccccccc@{}}
\toprule
\GraSLAMarkTop{liberopro}\multirow{2}{*}{\GraSLATableHeading{Method}} & \multirow{2}{*}{\GraSLATableHeading{Reference}} & \multirow{2}{*}{\GraSLATableHeading{PT.}} & \multirow{2}{*}{\GraSLATableHeading{Params (B) $\downarrow$}} & \multicolumn{5}{c}{\GraSLATableHeading{LIBERO-Pro Success Rate (\%) $\uparrow$}}\\
\cmidrule(l){5-9}
 & & & & \GraSLATableLabel{\textit{Object}} & \GraSLATableLabel{\textit{Position}} & \GraSLATableLabel{\textit{Semantic}} & \GraSLATableLabel{\textit{Task}} & \GraSLATableLabel{Avg.}\\
\midrule
\multicolumn{9}{@{}l}{\GraSLATableGroup{Large-scale}}\\
OpenVLA \cite{openvla,libero-pro} & CoRL~24 & \GraSLACheck & 7.54 & \GraSLARankTwo{93.00} & 0.00 & \GraSLARankTwo{97.25} & 0.00 & 47.56\\
$\pi_{0}$ \cite{pi0,libero-pro} & RSS~25 & \GraSLACheck & 3.20 & \GraSLARankThree{90.50} & 0.00 & 90.50 & 0.00 & 45.25\\
$\pi_{0.5}$ \cite{pi05,libero-pro} & CoRL~25 & \GraSLACheck & 3.35 & \GraSLARankOne{96.00} & \GraSLARankOne{20.75} & 95.75 & 0.75 & \GraSLARankTwo{53.31}\\
MolmoAct \cite{molmoact,libero-pro} & CoRL~25 & \GraSLACheck & 8.12 & 76.00 & 1.50 & 85.75 & 1.50 & 41.19\\
NORA \cite{nora,libero-pro} & arXiv~25 & \GraSLACheck & 3.76 & 70.50 & 0.00 & 86.25 & 0.00 & 39.19\\
GR00T-N1.7 \cite{grootn17,ectliberopro} & Report~26 & \GraSLACheck & 3.00 & 77.40 & 0.90 & 92.15 & 20.15 & 47.65\\
CLAP-4B (masking) \cite{clap} & CoRL~26 & \GraSLACross & 4.66 & 70.00 & 2.60 & 82.75 & \GraSLARankOne{24.00} & 44.84\\
JEPA-WAM \cite{jepawam,wavezero} & arXiv~26 & \GraSLACross & 1.35 & 68.20 & 0.05 & 81.50 & 20.55 & 42.58\\
Fast-WAM \cite{fastwam,wavezero} & NeurIPS~26 & \GraSLACross & 12.40 & 72.30 & 0.25 & 82.05 & 14.90 & 42.38\\
VLA-JEPA \cite{vlajepa,wavezero} & ECCV~26 & \GraSLACheck & 2.80 & 79.35 & 3.95 & 95.50 & \GraSLARankThree{20.65} & \GraSLARankThree{49.86}\\
\midrule
\multicolumn{9}{@{}l}{\GraSLATableGroup{Lightweight}}\\
X-VLA \cite{xvla,libero-pro} & ICLR~26 & \GraSLACheck & 0.88 & 79.00 & 0.75 & \GraSLARankThree{96.75} & 6.75 & 45.81\\
SmolVLA \cite{smolvla,predictivesupervision} & arXiv~25 & \GraSLACheck & \GraSLARankThree{0.45} & 47.30 & 0.00 & 22.10 & 13.90 & 20.83\\
SimVLA \cite{simvla} & arXiv~26 & \GraSLACross & 0.81 & 81.50 & \GraSLARankThree{8.25} & \GraSLARankOne{98.75} & 6.00 & 48.63\\
SUREFlow \cite{sureflow} & IROS~26 & \GraSLACross & \GraSLARankOne{0.179} & 68.50 & 0.00 & 87.75 & 0.00 & 39.06\\
VLA-0-0.8B \cite{vla0,clap} & arXiv~25 & \GraSLACross & 0.87 & 54.25 & 2.35 & 62.15 & 19.30 & 34.51\\
\GraSLAMarkBottom{liberopro}\GraSLATableHighlight \textit{GraSLA (ours)} & --- & \GraSLACross & \GraSLARankTwo{0.29} & 80.55 & \GraSLARankTwo{12.65} & 96.15 & \GraSLARankTwo{23.95} & \GraSLARankOne{53.33}\\
\bottomrule
\end{tabular*}
\end{table*}

\subsection{Experimental Setup}
\textbf{Training.} GraSLA employs Utonia \citep{utonia} as the spatial encoder, BERT-Base (uncased) \citep{bert} as the language encoder, and GSNet \citep{graspness} with a learnable GraspEncoder as the grasp branch. Only BERT and GSNet are frozen, while the remaining modules are trainable. Training details are provided in Appendix~\ref{app:training}.

\textbf{LIBERO.} Evaluation covers the \textit{Spatial}, \textit{Object}, \textit{Goal}, and \textit{Long} suites of LIBERO \citep{libero}, each containing ten language-conditioned tasks. The 3D training dataset is constructed by replaying demonstrations from the modified \texttt{no\_noops} RLDS data \citep{libero-modified}. Following prior work \citep{openvla,openvlaoft}, each task is evaluated over 50 rollouts, reporting the success rate for each suite and the overall average.

\textbf{LIBERO+.} The LIBERO-trained policy is evaluated without retraining on 10,030 task variants in LIBERO+ \citep{libero-plus}, spanning seven perturbation dimensions: \textit{Camera}, \textit{Robot}, \textit{Language}, \textit{Light}, \textit{Background}, \textit{Noise}, and \textit{Layout}. Following the official protocol, each variant is evaluated for one rollout, reporting the success rate for each dimension and overall.

\textbf{LIBERO-Pro.} The same policy is further evaluated on four perturbation types in LIBERO-Pro \citep{libero-pro}: \textit{Object}, \textit{Position}, \textit{Semantic}, and \textit{Task}. Following its protocol, each perturbed task is evaluated over 50 initial states, reporting the success rate for each type and their average.

\subsection{Main Results}
We compare GraSLA with representative large-scale and lightweight methods. Beyond task performance under standard settings, the comparisons also examine robustness to scene and instruction perturbations, along with the computational cost of these gains.

\textbf{LIBERO Results.} Table~\ref{tab:main} summarizes the results on LIBERO. Without robot-data pretraining, GraSLA achieves the highest average success rate of 99.3\% while remaining among the most lightweight and fastest methods at 0.29B parameters and 39.5 ms latency. The largest gains appear on the \textit{Goal} and \textit{Long} suites, where tasks either share identical scenes under different instructions or chain two subtasks across different objects. Both settings require identifying the target object at each stage. By binding grasp hypotheses to the scene and instruction via SLG interaction, GraSLA matches or surpasses models over an order of magnitude larger.

\textbf{LIBERO+ Results.} Under the broad perturbations of LIBERO+ in Table~\ref{tab:liberoplus-main}, GraSLA achieves the highest overall success rate of 73.7\%, with the widest margins on \textit{Camera} at 81.6\% and \textit{Noise} at 86.4\%. Since \textit{Camera} and \textit{Noise} alter third-person observations without changing 3D geometry, our multi-view point clouds and wrist-view grasp hypotheses remain anchored to object surfaces. In contrast, \textit{Light} and \textit{Background} perturbations alter point-cloud colors and weaken the spatial features extracted by Utonia, where GraSLA achieves 79.1\% and 77.4\%.

\textbf{LIBERO-Pro Results.} On LIBERO-Pro in Table~\ref{tab:liberopro-main}, GraSLA achieves the highest average success rate of 53.33\%, matching $\pi_{0.5}$ at 53.31\% while using roughly one-tenth of its parameters and without robot-data pretraining. Notably, most baselines collapse to near zero under \textit{Position} and \textit{Task} perturbations, where objects are relocated or the target changes, whereas GraSLA reaches 12.65\% on \textit{Position} and 23.95\% on \textit{Task}. Because grasp hypotheses are regenerated from current observations, they naturally follow relocated objects. SLG interaction then selects the hypotheses matching the new instruction, providing reliable guidance for action prediction.

\subsection{Ablation Studies}
To examine the role of different components, we conduct ablations on LIBERO, as shown in Table \ref{tab:libero-ablation}.

\textbf{Input Branches.} Omitting any of the $\mathbf S$, $\mathbf L$, or $\mathbf G$ branches lowers the average success rate, whereas \textit{Object} stays near saturation across all variants. Dropping $\mathbf L$ causes the steepest decline, reducing \textit{Goal} to near chance at 10.2\%, since tasks in \textit{Goal} share identical scenes and are distinguished solely by the instruction. Without $\mathbf S$, performance drops primarily on \textit{Long} at 59.8\% and \textit{Spatial} at 86.6\%, which depend on multi-stage localization and spatial relations. Finally, removing $\mathbf G$ lowers success rates across \textit{Spatial}, \textit{Goal}, and \textit{Long}, reducing the overall average from 99.30\% to 98.10\% and confirming that grasp hypotheses supply geometric priors complementary to spatial and language representations.

\textbf{Grasp Representation.} Ablating either the gripper configuration or the local features noticeably degrades \textit{Long}, dropping from 98.8\% to 92.6\% and 90.6\%, respectively. Both components are therefore essential for effective geometric priors, as the gripper configuration specifies pose constraints while local features capture surface geometry around each hypothesis.

\textbf{Multimodal Fusion.} Across fusion variants, simpler suites change only slightly, whereas \textit{Long} is most sensitive. Dropping $\operatorname{LE}$ or the SLG blocks causes the largest drops on \textit{Long}, to 89.6\% and 91.8\%. This shows that both cross-modal fusion and language refinement are essential for multi-stage tasks. Removing $\mathbf G\leftrightarrow\{\mathbf S,\mathbf L\}$ or individual grasp feedback paths also lowers \textit{Long}, and $\mathbf G\!\to\!\mathbf S$ causes the largest drop to 92.2\%, showing the importance of injecting approach geometry back into spatial representations.

\begin{table}[!t]
\centering\fontsize{9}{10.8}\selectfont
\captionsetup{name=Table}
\setlength{\tabcolsep}{1pt}
\renewcommand{\arraystretch}{1.10}
\caption{LIBERO ablations. Training follows the setup used for GraSLA in Table \ref{tab:main}. $\cdot\leftrightarrow\cdot$ denotes bidirectional interactions, and $\mathbf{G}\rightarrow\cdot$ denotes grasp feedback to $\mathbf{S}$, $\mathbf{L}$, or both. For fair comparison, we adjust module depths to closely match the number of parameters in the full model. The top three results in each column are shown in \GraSLARankOne{red}, \GraSLARankTwo{green}, and \GraSLARankThree{blue}.}
\label{tab:libero-ablation}
\newlength{\GraSLAAblationLabelWidth}
\settowidth{\GraSLAAblationLabelWidth}{\textit{Grasp Representation}\hspace{2pt}}
\begin{tabular*}{\columnwidth}{@{\extracolsep{\fill}}l!{\GraSLATableDivider}ccccc@{}}
\toprule
\GraSLAMarkTop{ablation}\multirow{2}{*}{\makebox[\GraSLAAblationLabelWidth][l]{\textbf{Variant}}} & \multicolumn{5}{c}{\textbf{LIBERO Success Rate (\%)} $\uparrow$}\\
\cmidrule(l){2-6}
 & \textit{Spatial} & \textit{Object} & \textit{Goal} & \textit{Long} & Avg.\\
\midrule
\multicolumn{6}{@{}l}{\textit{Input Branches}}\\
w/o spatial $\mathbf S$ & 86.6 & \GraSLARankThree{99.6} & 90.6 & 59.8 & 84.15\\
w/o language $\mathbf L$ & 87.0 & \GraSLARankTwo{99.8} & 10.2 & 78.0 & 68.75\\
w/o grasp $\mathbf G$ & 97.4 & \GraSLARankThree{99.6} & 98.4 & \GraSLARankThree{97.0} & 98.10\\
\midrule
\multicolumn{6}{@{}l}{\textit{Grasp Representation}}\\
w/o gripper config. & 98.4 & \GraSLARankOne{100.0} & 99.2 & 92.6 & 97.55\\
w/o local feat. & 97.0 & 99.4 & 99.2 & 90.6 & 96.55\\
\midrule
\multicolumn{6}{@{}l}{\textit{Multimodal Fusion}}\\
w/o $\mathbf{SLG}$ blocks & \GraSLARankThree{98.6} & 99.2 & 98.6 & 91.8 & 97.05\\
w/o $\mathbf S\leftrightarrow\mathbf L$ & \GraSLARankThree{98.6} & \GraSLARankOne{100.0} & \GraSLARankTwo{99.6} & 95.6 & \GraSLARankThree{98.45}\\
w/o $\mathbf G\leftrightarrow\{\mathbf S,\mathbf L\}$ & \GraSLARankTwo{98.8} & \GraSLARankOne{100.0} & \GraSLARankThree{99.4} & 93.0 & 97.80\\
w/o $\operatorname{LE}$ & 98.4 & \GraSLARankOne{100.0} & \GraSLARankThree{99.4} & 89.6 & 96.85\\
w/o $\mathbf G\!\to\!\mathbf S$ & 97.6 & \GraSLARankOne{100.0} & 99.0 & 92.2 & 97.20\\
w/o $\mathbf G\!\to\!\mathbf L$ & 98.2 & \GraSLARankOne{100.0} & 99.2 & \GraSLARankTwo{97.8} & \GraSLARankTwo{98.80}\\
w/o $\mathbf G\!\to\!\{\mathbf S,\mathbf L\}$ & 98.4 & \GraSLARankOne{100.0} & \GraSLARankThree{99.4} & 93.8 & 97.90\\
\midrule
\GraSLAMarkBottom{ablation}\GraSLAColumnHighlight \textit{GraSLA (ours)} & \GraSLARankOne{99.6} & 99.0 & \GraSLARankOne{99.8} & \GraSLARankOne{98.8} & \GraSLARankOne{99.3}\\
\bottomrule
\end{tabular*}
\end{table}

\section{Conclusion and Future Work}
This work introduces GraSLA, a lightweight spatial-language-action model for language-conditioned manipulation with grasp hypotheses as explicit geometric priors. While pose-based methods lack responsiveness during open-loop execution and action-based methods generalize poorly due to implicit behavior learning, GraSLA combines the strengths of both. Specifically, it retains diverse grasp hypotheses and uses SLG interaction blocks to bind task-relevant hypotheses to scene context and task semantics for closed-loop action prediction. Extensive evaluations on LIBERO, LIBERO+, and LIBERO-Pro demonstrate strong performance across standard, perturbed, and shifted conditions with low parameter and latency overhead. Nevertheless, the current design targets parallel-jaw manipulation, where grasp hypotheses are less informative for non-prehensile skills such as pushing and pressing. Future work will explore broader contact-rich priors to support dexterous and multi-fingered manipulation.

{\small
\bibliographystyle{templates/kiramei/template/abbrvnat}
\bibliography{grasplens_en}
}

\onecolumn
\raggedbottom
\begin{center}
{\Large\bfseries GraSLA: A Lightweight Spatial-Language-Action Model with Grasp Hypotheses as Geometric Priors\par}
\vspace{0.5em}
{\Large Supplementary Material\par}
\vspace{1.0em}
\end{center}
\appendix
\section{Model and Training Details}
\label{app:training}

\begingroup
\setlength{\columnsep}{8pt}
\setlength{\intextsep}{-1.2pt}
\begin{wraptable}{r}{0.43\textwidth}
\centering
\setlength{\tabcolsep}{1.75pt}
\renewcommand{\arraystretch}{0.94}
\captionsetup{skip=3pt}
\caption{Model and training configurations on LIBERO.}
\label{tab:app-model}
\label{tab:app-training}
\begin{tabular}{@{}llc@{}}
\toprule
 & Setting & LIBERO\\
\midrule
\multirow{6}{*}{\textit{Model}} & Input views & Third + wrist\\
 & Points per view & $8\,192$\\
 & Spatial tokens & 512\\
 & Grasp tokens & 256\\
 & Robot state dim. & 8\\
 & Action chunk & $12\times 7$\\
\midrule
\multirow{7}{*}{\textit{Training}} & Training data & OpenVLA RLDS\\
 & Optimizer & AdamW\\
 & Learning rate & $5\times 10^{-5}$\\
 & Effective batch & 128\\
 & Warmup steps & $10\,000$\\
 & Training budget & $80\,000$\\
 & EMA decay & 0.999\\
\bottomrule
\end{tabular}
\end{wraptable}

\textbf{Input Processing.}
RGB-D observations are back-projected into colored point clouds and transformed to a common coordinate frame.

\noindent\textbf{Grasp Hypotheses.}
GSNet generates grasp hypotheses and local features from wrist-view point clouds. After collision checking and nonmaximum suppression (NMS), hypotheses are sampled by geometric region and collision state. For LIBERO, foreground grouping uses 5\,mm voxels, a 25\,mm radius, and at least 20 occupied voxels per component. NMS uses position and rotation thresholds of 30\,mm and $\pi/6$.

\noindent\textbf{Model Configuration.}
Six SLG blocks fuse the 256-dimensional spatial, language, and grasp tokens. A three-layer decoder predicts action chunks from the fused tokens and robot state. Table~\ref{tab:app-model} lists the input and action configurations.

\noindent\textbf{Training Details.}
GraSLA uses a masked $\ell_1$ loss on normalized action chunks. AdamW uses $(\beta_1,\beta_2)=(0.9,0.95)$, $\epsilon=10^{-8}$, weight decay $10^{-10}$, and a gradient norm limit of 1. The learning rate warms up linearly from 10\% of its peak and remains constant. Table~\ref{tab:app-training} lists the settings.
\par\endgroup

\section{Additional Results}
\label{app:additional-results}

\begingroup
\setlength{\parindent}{0pt}
\setlength{\columnsep}{10pt}
\setlength{\intextsep}{-1.2pt}
\begin{wraptable}{r}{0.50\textwidth}
\raggedright
\setlength{\tabcolsep}{2.5pt}
\renewcommand{\arraystretch}{0.94}
\captionsetup{skip=3pt}
\caption{Per-task results of GraSLA on LIBERO, reported as successful/total rollouts. Task IDs follow the benchmark order.}
\label{tab:libero-task-results}
\begin{tabular*}{\linewidth}{@{\extracolsep{\fill}}lrrrr@{}}
\toprule
ID & \textit{Spatial} & \textit{Object} & \textit{Goal} & \textit{Long}\\
\midrule
0 & 50/50 & 50/50 & 50/50 & 48/50\\
1 & 49/50 & 50/50 & 50/50 & 50/50\\
2 & 50/50 & 50/50 & 50/50 & 50/50\\
3 & 50/50 & 50/50 & 49/50 & 50/50\\
4 & 50/50 & 50/50 & 50/50 & 50/50\\
5 & 49/50 & 45/50 & 50/50 & 50/50\\
6 & 50/50 & 50/50 & 50/50 & 50/50\\
7 & 50/50 & 50/50 & 50/50 & 50/50\\
8 & 50/50 & 50/50 & 50/50 & 46/50\\
9 & 50/50 & 50/50 & 50/50 & 50/50\\
\midrule
Total & 498/500 & 495/500 & 499/500 & 494/500\\
\bottomrule
\end{tabular*}
\end{wraptable}

To complement the main results, we provide task-level results on LIBERO, LIBERO+, and LIBERO-Pro.

\noindent\textbf{LIBERO.}
Table~\ref{tab:libero-task-results} presents results across the four suites. GraSLA succeeds in at least 45 of 50 rollouts on every task, including the long-horizon tasks in \textit{Long}.

\noindent\textbf{LIBERO+.}
Table~\ref{tab:liberoplus-task-results} reports results for each task across seven perturbation dimensions. GraSLA maintains high success on many tasks under \textit{Camera}, \textit{Light}, and \textit{Noise} perturbations, showing robustness to changes in visual observations.

\noindent\textbf{LIBERO-Pro.}
Table~\ref{tab:liberopro-task-results} presents results under the four perturbation types. Under \textit{Semantic} perturbations, GraSLA succeeds in at least 45 of 50 rollouts on 36 of the 40 tasks, demonstrating consistent performance across semantic variations.
\par\endgroup

\begin{table}[H]
\centering\fontsize{9}{10.8}\selectfont
\setlength{\tabcolsep}{2pt}
\renewcommand{\arraystretch}{1.05}
\caption{Per-task results on LIBERO+, reported as successful/total rollouts. Task IDs follow the benchmark order.}
\label{tab:liberoplus-task-results}
\begin{tabular*}{\textwidth}{@{\extracolsep{\fill}}llrrrrrrrr@{}}
\toprule
Suite & ID & \textit{Camera} & \textit{Robot} & \textit{Language} & \textit{Light} & \textit{Background} & \textit{Noise} & \textit{Layout} & Overall\\
\midrule
\multirow{10}{*}{\textit{Spatial}} & 0 & 6/6 & 18/32 & 42/46 & 15/42 & 18/29 & 31/32 & 32/44 & 162/231\\
 & 1 & 37/37 & 26/38 & 22/32 & 10/10 & 10/10 & 34/34 & 37/37 & 176/198\\
 & 2 & 34/35 & 29/35 & 28/28 & 50/50 & 46/46 & 35/35 & 38/38 & 260/267\\
 & 3 & 49/49 & 17/24 & 40/42 & 24/24 & 31/31 & 31/31 & 45/45 & 237/246\\
 & 4 & 47/48 & 11/40 & 14/37 & 44/44 & 26/26 & 40/40 & 40/45 & 222/280\\
 & 5 & 44/46 & 15/32 & 31/32 & 34/34 & 34/35 & 37/37 & 39/40 & 234/256\\
 & 6 & 42/42 & 20/27 & 33/46 & 31/31 & 20/20 & 34/34 & 31/32 & 211/232\\
 & 7 & 43/43 & 24/43 & 35/35 & 24/32 & 39/41 & 41/41 & 32/34 & 238/269\\
 & 8 & 33/33 & 30/37 & 44/45 & 4/4 & 6/6 & 28/28 & 26/27 & 171/180\\
 & 9 & 37/37 & 27/42 & 36/47 & 21/21 & 14/14 & 39/39 & 40/43 & 214/243\\
\midrule
\multirow{10}{*}{\textit{Object}} & 0 & 21/21 & 17/42 & 23/23 & 41/41 & 32/32 & 42/42 & 39/41 & 215/242\\
 & 1 & 20/20 & 16/39 & 29/34 & 4/4 & 4/5 & 44/44 & 24/27 & 141/173\\
 & 2 & 47/47 & 28/40 & 32/36 & 32/32 & 25/25 & 42/42 & 34/42 & 240/264\\
 & 3 & 0/47 & 14/38 & 10/26 & 5/49 & 1/47 & 28/41 & 31/44 & 89/292\\
 & 4 & 0/47 & 2/38 & 5/37 & 0/13 & 0/11 & 0/43 & 3/25 & 10/214\\
 & 5 & 50/50 & 19/38 & 28/36 & 22/22 & 13/13 & 44/44 & 44/48 & 220/251\\
 & 6 & 49/49 & 24/42 & 28/28 & 45/45 & 21/21 & 44/44 & 38/44 & 249/273\\
 & 7 & 48/48 & 19/40 & 41/42 & 26/26 & 32/32 & 39/39 & 41/45 & 246/272\\
 & 8 & 41/41 & 13/34 & 49/49 & 31/31 & 22/22 & 42/42 & 35/38 & 233/257\\
 & 9 & 26/26 & 26/47 & 30/43 & 34/34 & 40/40 & 41/41 & 40/49 & 237/280\\
\midrule
\multirow{10}{*}{\textit{Goal}} & 0 & 21/21 & 22/43 & 29/46 & 42/42 & 30/30 & 42/42 & 36/44 & 222/268\\
 & 1 & 32/32 & 20/38 & 8/19 & 4/4 & 6/6 & 29/29 & 10/28 & 109/156\\
 & 2 & 0/40 & 6/40 & 3/39 & 0/28 & 3/41 & 0/38 & 3/44 & 15/270\\
 & 3 & 0/48 & 0/45 & 0/50 & 0/50 & 1/45 & 0/41 & 0/45 & 1/324\\
 & 4 & 50/50 & 26/43 & 38/44 & 13/13 & 13/13 & 46/46 & 16/34 & 202/243\\
 & 5 & 0/49 & 0/40 & 0/39 & 0/36 & 0/27 & 0/33 & 1/46 & 1/270\\
 & 6 & 47/49 & 29/47 & 18/48 & 25/25 & 19/32 & 47/47 & 25/43 & 210/291\\
 & 7 & 46/46 & 20/37 & 4/40 & 21/21 & 11/11 & 29/29 & 48/48 & 179/232\\
 & 8 & 31/31 & 17/37 & 28/35 & 26/26 & 30/30 & 32/32 & 25/44 & 189/235\\
 & 9 & 42/42 & 34/39 & 25/50 & 34/34 & 46/46 & 42/42 & 40/49 & 263/302\\
\midrule
\multirow{10}{*}{\textit{Long}} & 0 & 20/20 & 19/41 & 24/45 & 13/13 & 13/13 & 46/46 & 34/34 & 169/212\\
 & 1 & 32/32 & 21/35 & 19/21 & 18/18 & 35/35 & 35/35 & 28/34 & 188/210\\
 & 2 & 47/47 & 35/44 & 40/41 & 42/42 & 33/33 & 48/48 & 15/26 & 260/281\\
 & 3 & 41/48 & 31/44 & 47/47 & 49/49 & 45/45 & 48/48 & 21/23 & 282/304\\
 & 4 & 0/47 & 0/38 & 0/42 & 0/33 & 0/44 & 0/48 & 0/35 & 0/287\\
 & 5 & 46/46 & 15/34 & 28/28 & 22/22 & 13/13 & 46/46 & 45/48 & 215/237\\
 & 6 & 46/46 & 22/42 & 46/46 & 31/31 & 34/39 & 43/44 & 34/35 & 256/283\\
 & 7 & 50/50 & 12/39 & 41/41 & 4/4 & 8/8 & 39/39 & 11/13 & 165/194\\
 & 8 & 40/43 & 16/34 & 30/30 & 24/24 & 31/31 & 46/46 & 13/25 & 200/233\\
 & 9 & 40/40 & 34/42 & 37/42 & 38/38 & 28/28 & 49/49 & 38/39 & 264/278\\
\bottomrule
\end{tabular*}
\end{table}

\begin{table}[H]
\centering\fontsize{9}{10.8}\selectfont
\setlength{\tabcolsep}{0.9pt}
\renewcommand{\arraystretch}{1.05}
\caption{Per-task results on LIBERO-Pro, reported as successful/total rollouts.}
\label{tab:liberopro-task-results}
\begin{tabular*}{\columnwidth}{@{\extracolsep{\fill}}ccccc@{\hspace{0.6em}}ccccc@{}}
\toprule
\multicolumn{5}{c}{\textit{Spatial}} & \multicolumn{5}{c}{\textit{Goal}}\\
\cmidrule(r){1-5}\cmidrule(l){6-10}
ID & {\textit{Object}} & {\textit{Position}} & {\textit{Semantic}} & {\textit{Task}} & ID & {\textit{Object}} & {\textit{Position}} & {\textit{Semantic}} & {\textit{Task}}\\
\midrule
0 & 47/50 & 45/50 & 47/50 & 0/50 & 0 & 50/50 & 0/50 & 50/50 & 0/50\\
1 & 50/50 & 4/50 & 50/50 & 50/50 & 1 & 50/50 & 0/50 & 50/50 & 0/50\\
2 & 50/50 & 47/50 & 50/50 & 0/50 & 2 & 0/50 & 0/50 & 50/50 & 0/50\\
3 & 50/50 & 0/50 & 50/50 & 0/50 & 3 & 37/50 & 19/50 & 48/50 & 0/50\\
4 & 36/50 & 20/50 & 47/50 & 0/50 & 4 & 2/50 & 0/50 & 50/50 & 0/50\\
5 & 50/50 & 28/50 & 49/50 & 50/50 & 5 & 50/50 & 0/50 & 50/50 & 13/50\\
6 & 50/50 & 0/50 & 50/50 & 50/50 & 6 & 40/50 & 0/50 & 50/50 & 0/50\\
7 & 50/50 & 0/50 & 50/50 & 29/50 & 7 & 50/50 & 15/50 & 50/50 & 50/50\\
8 & 48/50 & 19/50 & 45/50 & 24/50 & 8 & 50/50 & 0/50 & 50/50 & 0/50\\
9 & 50/50 & 0/50 & 35/50 & 50/50 & 9 & 50/50 & 0/50 & 50/50 & 0/50\\
\midrule
\multicolumn{5}{c}{\textit{Object}} & \multicolumn{5}{c}{\textit{Long}}\\
\cmidrule(r){1-5}\cmidrule(l){6-10}
ID & {\textit{Object}} & {\textit{Position}} & {\textit{Semantic}} & {\textit{Task}} & ID & {\textit{Object}} & {\textit{Position}} & {\textit{Semantic}} & {\textit{Task}}\\
\midrule
0 & 36/50 & 0/50 & 50/50 & 0/50 & 0 & 4/50 & 0/50 & 49/50 & 37/50\\
1 & 50/50 & 0/50 & 50/50 & 50/50 & 1 & 50/50 & 4/50 & 50/50 & 18/50\\
2 & 50/50 & 0/50 & 38/50 & 0/50 & 2 & 37/50 & 0/50 & 50/50 & 0/50\\
3 & 27/50 & 44/50 & 50/50 & 5/50 & 3 & 50/50 & 0/50 & 50/50 & 0/50\\
4 & 2/50 & 0/50 & 33/50 & 0/50 & 4 & 50/50 & 0/50 & 34/50 & 0/50\\
5 & 50/50 & 0/50 & 50/50 & 0/50 & 5 & 30/50 & 0/50 & 50/50 & 0/50\\
6 & 50/50 & 2/50 & 50/50 & 0/50 & 6 & 50/50 & 5/50 & 50/50 & 2/50\\
7 & 50/50 & 0/50 & 50/50 & 0/50 & 7 & 2/50 & 1/50 & 50/50 & 1/50\\
8 & 50/50 & 0/50 & 50/50 & 0/50 & 8 & 48/50 & 0/50 & 50/50 & 50/50\\
9 & 50/50 & 0/50 & 48/50 & 0/50 & 9 & 15/50 & 0/50 & 50/50 & 0/50\\
\bottomrule
\end{tabular*}
\end{table}

\section{Additional Ablations}
\label{app:additional-ablation}
The module ablations follow the training setup in Appendix~\ref{app:training}. Module depths are adjusted to closely match the full model's parameter count.

\subsection{LIBERO Ablations}

Tables~\ref{tab:app-encoding-modules} to~\ref{tab:app-execution-horizon} present component and hyperparameter ablations on standard LIBERO.

\noindent\textbf{Encoding Modules.} Table~\ref{tab:app-encoding-modules} compares alternative grasp detectors, spatial encoders, and language encoders. Our default configuration with GSNet, Utonia, and BERT achieves the highest average success rate of 99.30\% at 0.29B parameters. Replacing GSNet with EconomicGrasp reduces the average to 97.80\%, with the main drop on \textit{Long} from 98.8\% down to 94.4\%. Substituting Utonia with Sonata or Concerto lowers performance to 96.90\% and 94.40\%, respectively, showing that Utonia provides stronger geometric representations for multi-step manipulation. Replacing BERT with GTE-base or ModernBERT also lowers standard LIBERO accuracy to 97.70\% and 97.95\%, respectively, because BERT is pretrained purely with masked language modeling and preserves fine-grained token-level distinctions that fit seen task templates well.

\noindent\textbf{SLG Interaction Depth.} We vary the number of SLG interaction blocks $N$ from 2 to 8 in Table~\ref{tab:app-slg-depth}. Average performance peaks at $N=6$ with 99.30\%. While $N=8$ reaches 100.0\% on both \textit{Object} and \textit{Goal}, it drops slightly on \textit{Spatial} and \textit{Long} to a 99.15\% average and degrades zero-shot generalization on LIBERO+ to 70.4\% compared to 73.7\% in Table~\ref{tab:plus-ablation}, indicating that $N=6$ provides the best balance between fusion capacity and generalization.

\noindent\textbf{Action Execution Horizon.} We vary the execution horizon $h$ between policy predictions in Table~\ref{tab:app-execution-horizon}, where extra action queries for $h=15$ are linearly extrapolated from the last four learned queries. Performance peaks at $h=12$ with 99.30\% average success. A short horizon of $h=3$ fragments multi-step execution and lowers \textit{Long} by 7.0 percentage points from 98.8\% down to 91.8\%, whereas an overly long open-loop horizon of $h=15$ reduces closed-loop reactivity and drops the average to 97.85\%.

\begin{table}[!h]
\centering
\caption{LIBERO success rates (\%) with different encoding modules. The top three results in each column are shown in \GraSLARankOne{red}, \GraSLARankTwo{green}, and \GraSLARankThree{blue}.}
\label{tab:app-encoding-modules}
\begin{tabular*}{\linewidth}{@{\extracolsep{\fill}}lcccccc@{}}
\toprule
\textbf{Variant} & \textbf{Total Params (B)} $\downarrow$ & \textit{Spatial} & \textit{Object} & \textit{Goal} & \textit{Long} & Avg.\\
\midrule
\multicolumn{7}{@{}l}{\textit{Grasp}}\\
EconomicGrasp \cite{economicgrasp} & \GraSLARankTwo{0.29} & 97.2 & \GraSLARankOne{100.0} & \GraSLARankTwo{99.6} & \GraSLARankThree{94.4} & \GraSLARankThree{97.80}\\
\midrule
\multicolumn{7}{@{}l}{\textit{Spatial}}\\
Sonata \cite{sonata} & \GraSLARankOne{0.26} & 94.2 & \GraSLARankThree{99.6} & \GraSLARankTwo{99.6} & 94.2 & 96.90\\
Concerto \cite{concerto} & \GraSLARankOne{0.26} & 93.2 & \GraSLARankOne{100.0} & 98.8 & 85.6 & 94.40\\
\midrule
\multicolumn{7}{@{}l}{\textit{Language}}\\
GTE-base \cite{gte} & \GraSLARankThree{0.32} & \GraSLARankTwo{98.8} & \GraSLARankTwo{99.8} & \GraSLARankThree{99.0} & 93.2 & 97.70\\
ModernBERT \cite{modernbert} & 0.33 & \GraSLARankThree{97.8} & \GraSLARankOne{100.0} & 98.0 & \GraSLARankTwo{96.0} & \GraSLARankTwo{97.95}\\
\midrule
\GraSLAColumnHighlight \textit{GraSLA (ours)} & \GraSLARankTwo{0.29} & \GraSLARankOne{99.6} & 99.0 & \GraSLARankOne{99.8} & \GraSLARankOne{98.8} & \GraSLARankOne{99.3}\\
\bottomrule
\end{tabular*}
\end{table}

\begin{table}[!h]
\centering
\caption{LIBERO success rates (\%) for different SLG interaction depths.}
\label{tab:app-slg-depth}
\begin{tabular*}{\linewidth}{@{\extracolsep{\fill}}lccccc@{}}
\toprule
\textbf{$N$} & \textit{Spatial} & \textit{Object} & \textit{Goal} & \textit{Long} & Avg.\\
\midrule
$2$ & \GraSLARankTwo{98.8} & \GraSLARankOne{100.0} & \GraSLARankThree{99.2} & \GraSLARankThree{97.4} & \GraSLARankThree{98.85}\\
$4$ & \GraSLARankThree{98.4} & \GraSLARankOne{100.0} & 99.0 & 93.8 & 97.8\\
\GraSLAColumnHighlight $6$ & \GraSLARankOne{99.6} & \GraSLARankTwo{99.0} & \GraSLARankTwo{99.8} & \GraSLARankOne{98.8} & \GraSLARankOne{99.3}\\
$8$ & \GraSLARankTwo{98.8} & \GraSLARankOne{100.0} & \GraSLARankOne{100.0} & \GraSLARankTwo{97.8} & \GraSLARankTwo{99.15}\\
\bottomrule
\end{tabular*}
\end{table}

\begin{table}[!h]
\centering
\caption{LIBERO success rates (\%) for different action execution horizons.}
\label{tab:app-execution-horizon}
\begin{tabular*}{\linewidth}{@{\extracolsep{\fill}}lccccc@{}}
\toprule
\textbf{$h$} & \textit{Spatial} & \textit{Object} & \textit{Goal} & \textit{Long} & Avg.\\
\midrule
$3$ & \GraSLARankOne{99.8} & 98.2 & 96.6 & 91.8 & 96.6\\
$6$ & \GraSLARankThree{99.2} & \GraSLARankOne{99.8} & \GraSLARankTwo{99.2} & 94.6 & \GraSLARankThree{98.2}\\
$9$ & \GraSLARankThree{99.2} & \GraSLARankOne{99.8} & \GraSLARankThree{98.6} & \GraSLARankTwo{97.2} & \GraSLARankTwo{98.7}\\
\GraSLAColumnHighlight $12$ & \GraSLARankTwo{99.6} & \GraSLARankThree{99.0} & \GraSLARankOne{99.8} & \GraSLARankOne{98.8} & \GraSLARankOne{99.3}\\
$15$ & 99.0 & \GraSLARankTwo{99.2} & 97.4 & \GraSLARankThree{95.8} & 97.85\\
\bottomrule
\end{tabular*}
\end{table}

\FloatBarrier
\subsection{LIBERO+ Ablations}

Table~\ref{tab:plus-ablation} reports zero-shot ablations under LIBERO+ perturbations across input branches, grasp representations, fusion designs, encoding modules, and SLG interaction depth.

\noindent\textbf{Input, Grasp, and Fusion Components.} Removing $\mathbf S$ causes the steepest drop to 49.2\% overall and collapses \textit{Language} to 23.3\%, confirming that language grounding under perturbations relies on spatial context. Removing $\mathbf L$ or $\mathbf G$ reduces overall success to 60.0\% and 72.5\%, respectively, while omitting SLG interaction blocks drops it to 67.2\%. Notably, removing $\mathbf S\leftrightarrow\mathbf L$ yields 74.6\% overall due to higher \textit{Camera} and \textit{Light} scores, yet severely degrades \textit{Robot} perturbations from 49.9\% down to 38.7\%, confirming its role under initial-state shifts.

\noindent\textbf{Encoding and Depth Variants under Perturbations.} Consistent with LIBERO, Utonia at 73.7\% outperforms Sonata at 62.2\% and Concerto at 60.4\% by over 11 percentage points, and $N=6$ achieves the highest \textit{Layout} score at 74.2\% while outperforming $N=2$, $4$, and $8$ overall at 73.7\% compared to 72.1\%, 71.5\%, and 70.4\%. Among language encoders, GTE-base achieves the highest overall success rate at 75.4\% and strongest \textit{Language} robustness at 71.2\%, compared to 69.3\% for BERT and 68.8\% for ModernBERT. Unlike masked-language-modeling encoders, GTE-base is trained with multi-stage contrastive learning over text pairs, mapping paraphrased or reworded instructions closer together in embedding space and thus improving robustness to LIBERO+ language variations.

\begin{table}[H]
\centering
\caption{Zero-shot LIBERO+ ablations (success rates in \%). The top three results in each column are shown in \GraSLARankOne{red}, \GraSLARankTwo{green}, and \GraSLARankThree{blue}.}
\label{tab:plus-ablation}
\begin{tabular*}{\linewidth}{@{\extracolsep{\fill}}lcccccccc@{}}
\toprule
\textbf{Variant} & \textit{Camera} & \textit{Robot} & \textit{Language} & \textit{Light} & \textit{Background} & \textit{Noise} & \textit{Layout} & \textbf{Overall}\\
\midrule
\multicolumn{9}{@{}l}{\textit{Input Branches}}\\
w/o spatial $\mathbf S$ & 62.4 & 32.3 & 23.3 & 60.1 & 58.8 & 64.3 & 47.9 & 49.2\\
w/o language $\mathbf L$ & 64.2 & 32.9 & 64.0 & 71.3 & 66.8 & 66.4 & 59.0 & 60.0\\
w/o grasp $\mathbf G$ & 83.7 & 46.5 & 66.6 & 79.6 & 77.0 & 85.4 & \GraSLARankThree{71.3} & 72.5\\
\midrule
\multicolumn{9}{@{}l}{\textit{Grasp Representation}}\\
w/o gripper config. & 80.2 & 46.6 & 62.8 & 79.9 & 77.8 & 83.8 & 68.9 & 70.9\\
w/o local feat. & 78.5 & 50.7 & 63.3 & 77.3 & 75.1 & 83.2 & 66.4 & 70.3\\
\midrule
\multicolumn{9}{@{}l}{\textit{Multimodal Fusion}}\\
w/o $\mathbf{SLG}$ blocks & 80.7 & 44.1 & 47.6 & 76.6 & 74.6 & 82.6 & 67.7 & 67.2\\
w/o $\mathbf S\leftrightarrow\mathbf L$ & \GraSLARankTwo{86.0} & 38.7 & \GraSLARankOne{73.8} & \GraSLARankOne{86.5} & \GraSLARankTwo{81.1} & \GraSLARankOne{87.8} & \GraSLARankTwo{72.6} & \GraSLARankTwo{74.6}\\
w/o $\mathbf G\leftrightarrow\{\mathbf S,\mathbf L\}$ & 83.1 & \GraSLARankOne{53.4} & 58.0 & \GraSLARankThree{81.4} & 78.6 & 84.3 & 67.6 & 71.8\\
w/o $\operatorname{LE}$ & 77.5 & \GraSLARankThree{51.4} & 60.2 & 77.0 & 76.6 & 82.7 & 67.3 & 69.9\\
w/o $\mathbf G\!\to\!\mathbf S$ & 82.6 & \GraSLARankTwo{52.6} & 58.2 & 79.2 & 79.1 & 83.5 & 70.0 & 71.7\\
w/o $\mathbf G\!\to\!\mathbf L$ & 82.9 & 42.9 & 61.3 & 78.1 & 78.3 & 84.5 & 66.6 & 70.1\\
w/o $\mathbf G\!\to\!\{\mathbf S,\mathbf L\}$ & 78.6 & 50.3 & 58.9 & 78.1 & 78.2 & 81.2 & 68.1 & 69.9\\
\midrule
\multicolumn{9}{@{}l}{\textit{Encoding Modules}}\\
EconomicGrasp \cite{economicgrasp} & 81.2 & 45.8 & 60.6 & 80.4 & 76.7 & 84.8 & 68.5 & 70.7\\
Sonata \cite{sonata} & 76.2 & 43.0 & 42.9 & 70.8 & 68.8 & 77.8 & 59.5 & 62.2\\
Concerto \cite{concerto} & 74.4 & 39.2 & 42.1 & 72.8 & 63.6 & 77.5 & 56.3 & 60.4\\
GTE-base \cite{gte} & \GraSLARankOne{87.3} & 48.6 & \GraSLARankTwo{71.2} & \GraSLARankTwo{85.2} & \GraSLARankOne{81.6} & \GraSLARankOne{87.8} & 69.9 & \GraSLARankOne{75.4}\\
ModernBERT \cite{modernbert} & \GraSLARankThree{84.0} & 45.0 & 68.8 & 79.9 & \GraSLARankThree{80.3} & \GraSLARankTwo{87.0} & 70.4 & 73.2\\
\midrule
\multicolumn{9}{@{}l}{\textit{SLG Interaction Depth}}\\
$N=2$ & 82.4 & 44.5 & \GraSLARankThree{69.3} & 80.0 & 75.6 & 84.6 & 70.6 & 72.1\\
$N=4$ & 81.5 & 48.2 & 67.2 & 77.7 & 76.0 & 83.8 & 68.2 & 71.5\\
$N=8$ & 80.4 & 49.3 & 64.9 & 77.9 & 72.3 & 81.0 & 68.7 & 70.4\\
\midrule
\GraSLAColumnHighlight \textit{GraSLA (ours)} & 81.6 & 49.9 & \GraSLARankThree{69.3} & 79.1 & 77.4 & \GraSLARankThree{86.4} & \GraSLARankOne{74.2} & \GraSLARankThree{73.7}\\
\bottomrule
\end{tabular*}
\end{table}

\subsection{LIBERO-Pro Ablations}

Table~\ref{tab:liberopro-ablation} reports zero-shot ablations under LIBERO-Pro perturbations. Across all variants, full GraSLA achieves the highest overall success rate at 53.33\%, as well as the best performance on \textit{Position} at 12.65\%, \textit{Semantic} at 96.15\%, and \textit{Task} at 23.95\%.

\textbf{Input, Grasp, and Fusion Components.} Removing $\mathbf S$ causes the steepest drop to 29.75\% overall and collapses \textit{Position} to 0.00\%, showing that spatial point-cloud features are indispensable when object locations change. Removing $\mathbf L$ reduces overall success to 36.60\% and collapses \textit{Task} to 3.55\%, as \textit{Task} perturbations require distinguishing modified instructions within similar scenes. Removing $\mathbf G$ drops overall success by 2.34 percentage points to 50.99\%, with clear declines on \textit{Position} from 12.65\% down to 7.75\% and \textit{Task} from 23.95\% down to 21.25\%, confirming that explicit grasp hypotheses improve generalization when objects relocate or instructions change. Dropping either the gripper configuration or local features lowers overall success further to 46.84\% and 46.61\%, showing that partial grasp encoding is less effective than a complete representation combining pose and surface context. Among fusion designs, removing SLG interaction blocks drops overall success to 41.28\%, while removing $\mathbf G\!\to\!\mathbf S$, $\operatorname{LE}$, or $\mathbf S\leftrightarrow\mathbf L$ degrades overall success to 43.04\%, 42.90\%, and 49.01\%, respectively, with $\mathbf S\leftrightarrow\mathbf L$ being particularly important for \textit{Task} at 14.15\% versus 23.95\%.

\textbf{Encoding and Depth Variants.} Consistent with LIBERO and LIBERO+, Utonia at 53.33\% substantially outperforms Sonata at 41.03\% and Concerto at 37.53\%, and GSNet outperforms EconomicGrasp at 44.84\%. Replacing BERT with GTE-base or ModernBERT lowers overall success to 48.83\% and 48.36\%, respectively, with sharp drops on \textit{Position} to 3.90\% and 3.05\% and on \textit{Semantic} to 88.35\% and 85.80\%, indicating that token-level masked language modeling in BERT pairs more reliably with spatial and grasp representations under object relocation and instruction shifts. Finally, varying SLG interaction depth shows that $N=6$ at 53.33\% outperforms $N=2$ at 48.43\%, $N=4$ at 50.20\%, and $N=8$ at 49.99\%, confirming the same optimal depth across all three benchmarks.

\begingroup
\setlength{\intextsep}{6pt}
\renewcommand{\arraystretch}{0.92}
\begin{table}[H]
\centering
\caption{Zero-shot LIBERO-Pro ablations (success rates in \%). The top three results in each column are shown in \GraSLARankOne{red}, \GraSLARankTwo{green}, and \GraSLARankThree{blue}.}
\label{tab:liberopro-ablation}
\begin{tabular*}{\linewidth}{@{\extracolsep{\fill}}lccccc@{}}
\toprule
\textbf{Variant} & \textit{Object} & \textit{Position} & \textit{Semantic} & \textit{Task} & \textbf{Overall}\\
\midrule
\multicolumn{6}{@{}l}{\textit{Input Branches}}\\
w/o spatial $\mathbf S$ & 54.75 & 0.00 & 52.75 & 11.50 & 29.75\\
w/o language $\mathbf L$ & 62.15 & \GraSLARankTwo{11.00} & 69.70 & 3.55 & 36.60\\
w/o grasp $\mathbf G$ & 79.05 & 7.75 & \GraSLARankTwo{95.90} & 21.25 & \GraSLARankTwo{50.99}\\
\midrule
\multicolumn{6}{@{}l}{\textit{Grasp Representation}}\\
w/o gripper config. & 75.50 & 2.85 & 86.20 & \GraSLARankThree{22.80} & 46.84\\
w/o local feat. & 72.30 & 7.15 & 83.65 & \GraSLARankTwo{23.35} & 46.61\\
\midrule
\multicolumn{6}{@{}l}{\textit{Multimodal Fusion}}\\
w/o $\mathbf{SLG}$ blocks & 72.60 & 5.60 & 73.20 & 13.70 & 41.28\\
w/o $\mathbf S\leftrightarrow\mathbf L$ & \GraSLARankOne{83.50} & 7.35 & 91.05 & 14.15 & 49.01\\
w/o $\mathbf G\leftrightarrow\{\mathbf S,\mathbf L\}$ & 80.30 & 3.15 & 79.90 & 20.55 & 45.98\\
w/o $\operatorname{LE}$ & 70.40 & 6.35 & 78.45 & 16.40 & 42.90\\
w/o $\mathbf G\!\to\!\mathbf S$ & 69.15 & 5.20 & 82.20 & 15.60 & 43.04\\
w/o $\mathbf G\!\to\!\mathbf L$ & 77.90 & 2.70 & 84.80 & 20.20 & 46.40\\
w/o $\mathbf G\!\to\!\{\mathbf S,\mathbf L\}$ & 77.95 & \GraSLARankThree{8.70} & 80.95 & 22.20 & 47.45\\
\midrule
\multicolumn{6}{@{}l}{\textit{Encoding Modules}}\\
EconomicGrasp \cite{economicgrasp} & 68.55 & 7.70 & 85.70 & 17.40 & 44.84\\
Sonata \cite{sonata} & 70.50 & 2.70 & 71.00 & 19.90 & 41.03\\
Concerto \cite{concerto} & 65.45 & 1.70 & 65.20 & 17.75 & 37.53\\
GTE-base \cite{gte} & \GraSLARankTwo{83.45} & 3.90 & 88.35 & 19.60 & 48.83\\
ModernBERT \cite{modernbert} & \GraSLARankThree{82.15} & 3.05 & 85.80 & 22.45 & 48.36\\
\midrule
\multicolumn{6}{@{}l}{\textit{SLG Interaction Depth}}\\
$N=2$ & 78.45 & 3.85 & 90.35 & 21.05 & 48.43\\
$N=4$ & 80.30 & 7.30 & \GraSLARankThree{91.50} & 21.70 & \GraSLARankThree{50.20}\\
$N=8$ & 81.50 & 9.30 & 88.10 & 21.05 & 49.99\\
\midrule
\GraSLAColumnHighlight \textit{GraSLA (ours)} & 80.55 & \GraSLARankOne{12.65} & \GraSLARankOne{96.15} & \GraSLARankOne{23.95} & \GraSLARankOne{53.33}\\
\bottomrule
\end{tabular*}
\end{table}
\endgroup

\section{Inference Latency Measurement}
\label{app:latency}
We measure inference latency on one NVIDIA RTX 5090 with batch size one, from CPU-resident observations to the complete CPU action output. The GPU is synchronized before and after each measurement. For each of the 40 LIBERO tasks $t$, five initial states define episodes $e$, and up to ten observations $o$ are sampled evenly from each episode, forming $\mathcal O=\{(t,e,o)\}$. Each run starts in a fresh process and uses ten held-out observations for warmup. Algorithm~\ref{alg:latency-measurement} summarizes the measurement and aggregation.

\begin{algorithm}[H]
\caption{Inference latency measurement}
\label{alg:latency-measurement}
\begin{algorithmic}[1]
\Require Policy $\pi$, observations $\mathcal O$, runs $R$
\Ensure Mean inference latency $\bar{\tau}$
\For{$r=1,\ldots,R$}
    \State Initialize and warm up $\pi$ in a fresh process
    \State Measure $\tau_{r,t,e,o}$ for each $(t,e,o)\in\mathcal O$
\EndFor
\State $\bar{\tau}_{r,t,e}\gets\operatorname{mean}_{o}\tau_{r,t,e,o}$
\State \Return $\operatorname{mean}_{r,t,e}\bar{\tau}_{r,t,e}$
\end{algorithmic}
\end{algorithm}

\clearpage

\section{Qualitative Rollouts}
\label{app:qualitative-rollouts}
\begin{figure}[H]
\centering
\includegraphics[width=\textwidth]{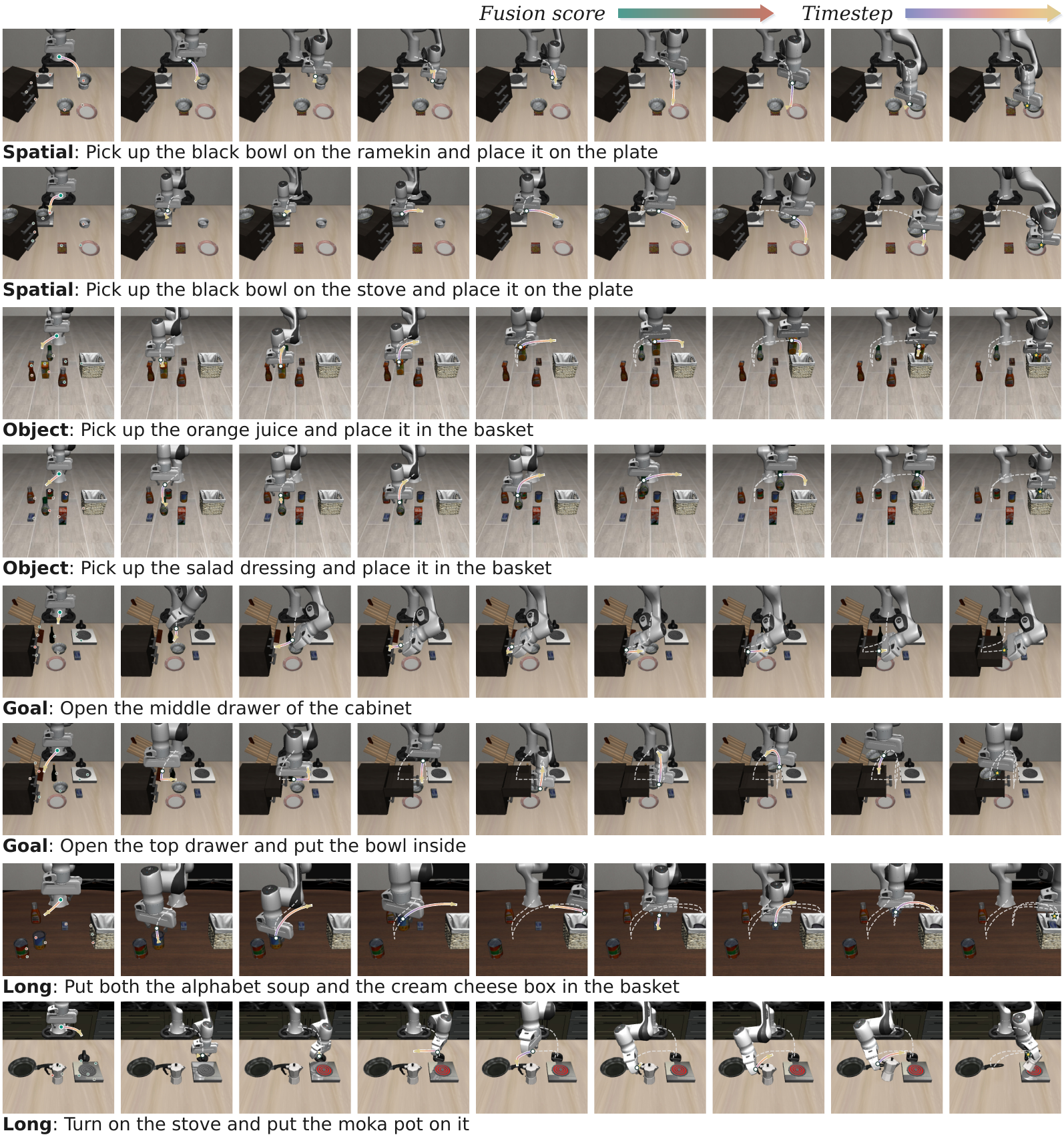}
\caption{Qualitative rollouts on LIBERO. Dots in the initial frame mark the centers of grasp hypotheses selected by NMS, with colors indicating their relative fusion attention scores. Solid curves show the next 24 executed end-effector steps, colored by timestep.}
\label{fig:qualitative-rollouts}
\end{figure}

\begin{figure}[H]
\centering
\includegraphics[width=\textwidth]{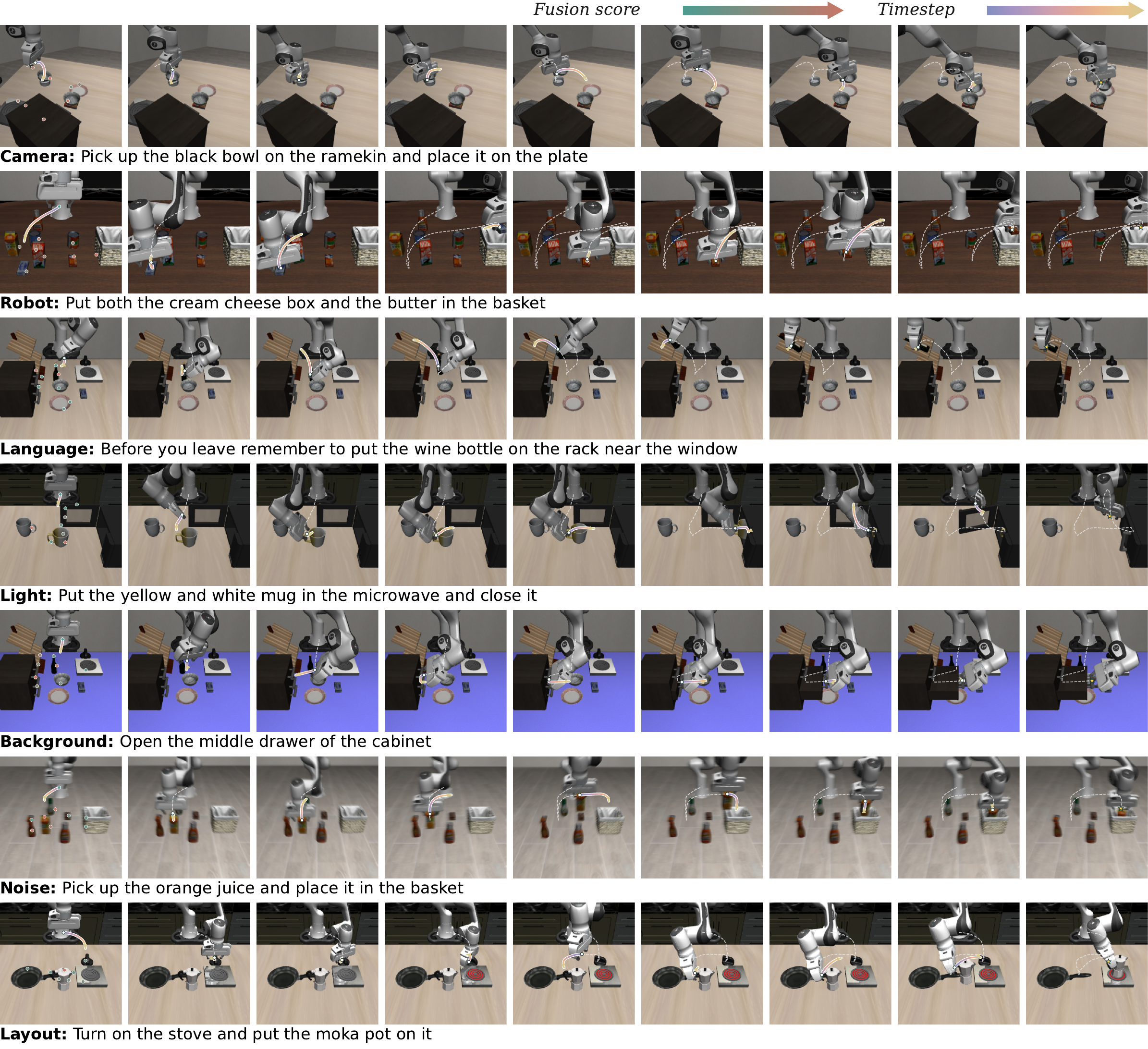}
\caption{Qualitative rollouts on LIBERO+ across seven perturbation dimensions. Dots in the first displayed frame mark the centers of grasp hypotheses selected by NMS, colored by their relative fusion attention scores. Solid curves show the next 24 executed end-effector steps, colored by timestep.}
\label{fig:qualitative-rollouts-liberoplus}
\end{figure}

\begin{figure}[H]
\centering
\includegraphics[width=\textwidth]{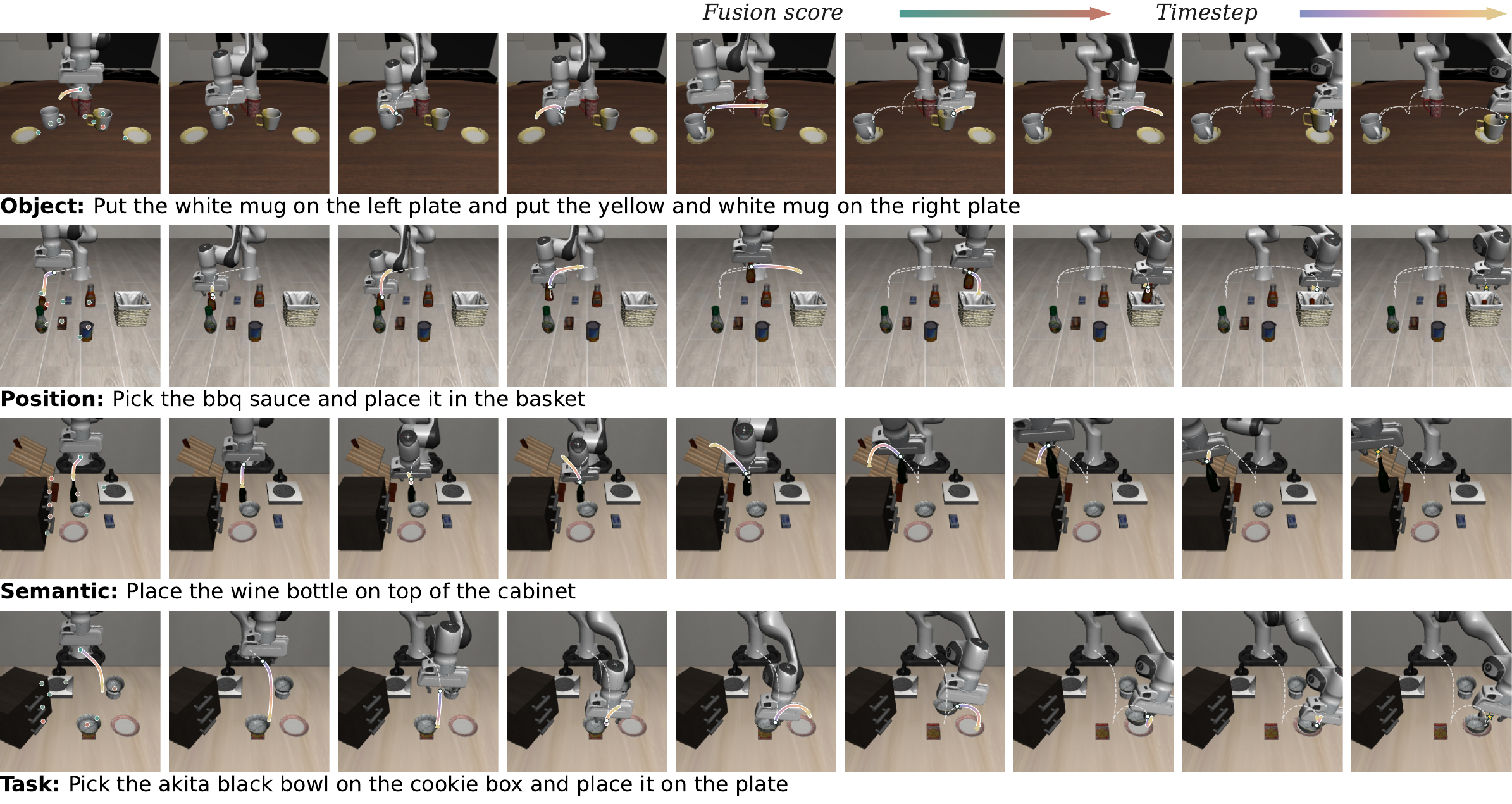}
\caption{Qualitative rollouts on LIBERO-Pro under four perturbation types. Dots and curves follow the conventions in Figure~\ref{fig:qualitative-rollouts-liberoplus}. The \textit{Object} example uses a wider rendering view to show both placement regions.}
\label{fig:qualitative-rollouts-liberopro}
\end{figure}

Figures~\ref{fig:qualitative-rollouts}--\ref{fig:qualitative-rollouts-liberopro} show rollouts on LIBERO, LIBERO+, and LIBERO-Pro. Grasp hypotheses provide geometric priors for subsequent action prediction.

\end{document}